%% file: main.tex
\documentclass[twoside,11pt]{article}
\usepackage{jair, theapa, rawfonts}

\usepackage{color}
\usepackage{xcolor}
\usepackage{overpic}
\usepackage{wrapfig}
\usepackage{subcaption}
\usepackage{graphicx}
\usepackage{amsmath}
\usepackage{amsfonts}
\usepackage{amsthm}
\usepackage{xspace}
\usepackage{colortbl}
\usepackage{booktabs}
\usepackage{multirow}
\usepackage[obeyFinal]{todonotes}
\usepackage{standalone}

\usetikzlibrary{patterns,backgrounds,fit}

\usepackage{url}
\usepackage{verbatim}
\usepackage{algorithm}
\usepackage{algpseudocode}
\usepackage{colortbl} 

\newcommand{\ALT}{{\texttt{ALTERNATE}}\xspace}
\newcommand{\SWI}{{\texttt{SWITCH}}\xspace}
\newcommand{\PRO}{{\texttt{TEXTUAL}}\xspace}
\newcommand{\UNE}{{\texttt{UNET}}\xspace}

\definecolor{ourred}{RGB}{252,0,0}
\definecolor{ourblue}{RGB}{0,0,252}

\jairheading{1}{2025}{1-15}{6/91}{9/91}
\ShortHeadings{Blending Concepts with Text-to-Image Diffusion Models}
{Olearo, Longari, Raganato, Pe\~naloza \& Melzi}
\firstpageno{1}

\begin{document}

\title{Blending Concepts with Text-to-Image Diffusion Models}

\author{\name Lorenzo Olearo \email lorenzo.olearo@unimib.it \\
	\name Giorgio Longari \email giorgio.longari@unimib.it  \\
	\name Alessandro Raganato \email alessandro.raganato@unimib.it \\
	\name Rafael Pe\~naloza \email rafael.penaloza@unimib.it \\
	\name Simone Melzi \email simone.melzi@unimib.it \\
	\addr University of Milano-Bicocca, Milan, Italy
}


\maketitle
\begin{abstract}
	Diffusion models have dramatically advanced text-to-image generation in recent years, translating abstract concepts into high-fidelity images with remarkable ease. In this work, we examine whether they can also \textit{blend} distinct concepts, ranging from concrete objects to intangible ideas, into coherent new visual entities under a zero-shot framework. Specifically, \textit{concept blending} merges the key attributes of multiple concepts (expressed as textual prompts) into a single, novel image that captures the essence of each concept. We investigate four blending methods, each exploiting different aspects of the diffusion pipeline (e.g., prompt scheduling, embedding interpolation, or layer-wise conditioning). Through systematic experimentation across diverse concept categories, such as merging concrete concepts, synthesizing compound words, transferring artistic styles, and blending architectural landmarks, we show that modern diffusion models indeed exhibit creative blending capabilities without further training or fine-tuning. Our extensive user study, involving 100 participants, reveals that no single approach dominates in all scenarios: each blending technique excels under certain conditions, with factors like prompt ordering, conceptual distance, and random seed affecting the outcome. These findings highlight the remarkable compositional potential of diffusion models while exposing their sensitivity to seemingly minor input variations.
\end{abstract}

\section{Introduction}
\label{sec:introduction}
\input{sections/introduction.tex}

\section{Related Work}
\label{sec:related-work}
\input{sections/related-work.tex}

\section{Preliminaries}
\label{sec:preliminaries}
\input{sections/preliminaries.tex}

\section{Blending Methods}
\label{sec:blending-methods}
\input{sections/blending-methods}

\section{Dependencies in Concept Blending Procedures}
\label{sec:analysis}
\input{sections/analysis}

\section{From Tangible to Intangible: Reimagining Conceptual Blending}
\label{sec:intangible}
\input{sections/intangible}

\newpage
\section{Validation and Results}
\label{sec:validation-and-results}
\input{sections/validation-and-results}

\section{Conclusions}
\label{sec:conclusions}
\input{sections/conclusions}

\vskip 0.2in
\bibliography{references}
\bibliographystyle{theapa}

\appendix
\section*{Appendix A. Additional Results}
\label{sec:appendix}
\input{sections/appendix}

\end{document}

%% file: sections/introduction.tex
Human cognition is grounded in our ability to form, manipulate, and recombine concepts—abstract representations that help us interpret and navigate the world. This cognitive flexibility underlies our creativity, allowing us to generate novel ideas by combining (or \emph{blending}) existing concepts in unexpected or previously unseen ways. Thus, even from a limited number of ``base'' concepts, it is possible to produce arbitrarily many new constructs.
The idea of concept blending, which originates from a theory of
cognition \cite{FaTu-08}, has recently been studied in more detail within the areas of knowledge representation, computational creativity, and others \cite{confalonieri2018concept}, using elements of conceptual spaces \cite{Gard04}, generalisation and specialisation \cite{CESK-AMAI18}, feature combination \cite{Porello2019-PORATO-2},
and other symbolic transformations.
As generative AI advances, and in particular text-to-image diffusion models, examining these architectures' ability to combine distinct concepts into a cohesive visual output without relying on specialized prompts, additional fine-tuning, or explicit symbolic engineering becomes increasingly important. 

This paper presents several methods on how the learned representations of these models, under a zero-shot paradigm, can unify multiple ideas into a single coherent image, showing their degree of creative synthesis that can arise in diffusion-based pipelines without further intervention. 
We investigate the capabilities of diffusion-based
text-to-image generative AI models to produce compelling images representing the blend of two concepts. Specifically, to generate visual syntheses of two or more concepts or ideas into one coherent element, relying only on their learned latent representations. In
other words, we examine: (i)~whether these models are able to produce blends as a zero-shot task, without additional training, fine-tuning, or prompting strategies; (ii)~how effective are different blending strategies based on the diffusion architecture to yield coherent and visually meaningful combinations; and (iii)~what is the role of factors like the conceptual distance, random seed selection, and specificity of the prompts employed in the ultimate success of the produced blend.
Through this lens, we highlight important limitations in the compositional capacities of diffusion models, even as we uncover the surprising potential for creative synthesis that emerges from their training.
Our work adopts text-to-image diffusion models \cite{dm-2015,dm-2020,stable-diffusion} as a testbed for this phenomenon by treating each concept as a (typically single-word) textual prompt, for instance, ``lion'' or ``cat'', and leveraging the model to produce their visual representations. By doing so, we bridge the gap between mental concept representations and their image-based instantiations, examining whether the recent generative pipelines capture a form of compositional creativity akin to what can be observed in human cognition. To our knowledge, this is the first comprehensive investigation that systematically compares multiple zero-shot blending techniques using a user-study–backed analysis, relying purely on the learned latent structure of large diffusion models.

In our recent work \cite{melzi2023does,olearo2024blend}, we proposed a preliminary approach to concept blending in diffusion models, showing initial evidence of their zero-shot blending capabilities. The present paper significantly expands that investigation in several key directions: (i) we clearly describe and motivate the blending methods employed, providing a comprehensive analysis of their fundamental properties; (ii) we introduce and motivate new categories of blends, broadening our experimental scenarios; (iii) we enhance our user study with a larger participant base and additional evaluation metrics; (iv) we extend the blending framework to multiple new domains beyond straightforward tangible prompts; and (v) we present novel insights into the role of seed variability in the diffusion pipeline.

To provide a thorough account, we conducted a user study with 100 participants who evaluated 22 concept pairs across four categories (namely, same class, different classes, compound words, choice of style, and architectural landmarks), controlling for both prompt ordering and random seeds. This systematic evaluation reveals important patterns around which methods excel in specific scenarios and how sensitive the blending process can be to seemingly minor input changes. 
Our main contributions are fourfold:
\begin{enumerate}
    \item \textbf{Comprehensive Comparison of Blending Methods.}
We present four zero-shot strategies, ranging from simple interpolation in the text embedding space to different U-Net conditioning, and detail each method's motivations, implementation, and practical trade-offs.

\item \textbf{Systematic Analysis of Dependencies.}
We show how the quality of a blend depends on factors including prompt design and random seed selection. Our results reveal patterns, biases, and constraints in how diffusion models represent composite concepts.

\item \textbf{Novel Blending Tasks and Explorations.}
We extend concept blending to broader scenarios, such as combining a specific painting with a general concept, or merging a concept with an emotion or style, evaluating diffusion models well beyond concrete object-object combinations.

\item \textbf{Empirical Findings and User Study.}
We validate each approach through both qualitative illustrations and comparative user feedback. Our evaluation highlights differences between methods, offering insights into when (and why) certain blends are more successful.
\end{enumerate}

Our experiments show that, although diffusion models can indeed generate compelling conceptual blends in certain cases, no single method universally outperforms the others, although some tend to behave better in general. 
Some strategies excel at merging highly dissimilar prompts, while others produce subtler merges for closely related concepts. Factors like the chosen seed or prompt ordering critically influence outcomes, revealing that concept blending in text-to-image systems remains sensitive and, at times, unpredictable; although this could be seen as a positive from a creativity point of view. 
Despite these limitations, we show that zero-shot blending, without fine-tuning, can be harnessed to produce imaginative visual hybrids, indicating a degree of embedded creativity in modern diffusion pipelines. Beyond its relevance for understanding how these models encode compositional knowledge, such blending capabilities could prove valuable for downstream AI tasks like creative design support or interactive brainstorming tools, which rely on the dynamic combination of concepts to spark new ideas \cite{sun2025creative}.

While our primary focus is on single-word prompts for clarity and tractability, we acknowledge that more complex or multi-word descriptions may further challenge (or improve) blending outcomes. We also emphasize that, while our methods systematically produce blended concepts, they do not replicate human conceptual integration. The deeper cognitive processes underlying genuine creativity remain an open area of research, and it is not our claim that these techniques capture or explain them fully. Nevertheless, by investigating diffusion-based blends in a controlled setting, we offer insights into both the potential and inherent constraints of state-of-the-art generative models as tools for AI-driven creativity. To ensure reproducibility, we have made code, user-study materials, and prompt configurations publicly available on GitHub.\footnote{\url{https://github.com/LorenzoOlearo/blending-diffusion-models}}  


The remainder of this paper is organized as follows. Section~\ref{sec:related-work} reviews related work on concept blending from both logical and visual perspectives followed, in Section~\ref{sec:preliminaries}, by a technical overview of diffusion models. 
In Section~\ref{sec:blending-methods}, we describe our experimental setup and introduce the four zero-shot blending methods that we analyse. Section~\ref{sec:analysis} then presents a detailed analysis of blending dependencies and limitations, followed by Section~\ref{sec:intangible}, where we reformulate the blending task to include styles and emotions. Section~\ref{sec:validation-and-results} discusses our user study methodology, results, and key findings. Section~\ref{sec:conclusions} concludes with a summary of contributions and future directions in this rapidly evolving field.

%% file: sections/related-work.tex




We first briefly recall the main work in the area of conceptual blending in general, before focusing on the specifics of visual blending.

\paragraph{Conceptual Blending}
Conceptual blending arose as a theory of cognition which proposes that element from diverse scenarios are continuously blended in our subconscious \cite{fauconnier2003conceptual}.
In the context of Artificial Intelligence, the task is often interpreted as trying to ``invent'' new concepts by combining the 
identifying features or properties of other existing concepts. A
thorough study of this idea in different knowledge domains is available in \cite{confalonieri2018concept}.

Since the original description based on mental spaces \cite{FaTu-CS98} is not fully specified, different approaches have been proposed, targeting specific 
characteristics of the spaces or the resulting blend 
\cite{blending-framework,kutz2014pluribus}. For instance, methods based on image schemas \cite{HEDBLOM201642} or description logics \cite{Porello2019-PORATO-2}.

The task of concept blending is also closely connected to that of building \emph{analogies} between concepts, in that one must 
understand the commonalities and differences between the underlying concepts. Results in analogical reasoning through word embeddings 
\cite{si2022word} suggests that embedding spaces behave, up to a point like conceptual spaces \emph{à la} Gardenfors \cite{Gard04}. Hence, it
is reasonable to explore the possibility of constructing blends through an exploration of embedding spaces. Yet, concept blends are
difficult to produce and explain verbally. A more natural approach is to build these concepts visually; specially if they are blends of two
concrete concepts. This brings us to the task of \emph{visual} concept blending.

\paragraph{Visual Blending}
An early contribution to the field of visual blending is presented in~\cite{ge2021visual}, where
the authors employed Large Language Models (LLMs) to generate prompts that
semantically blend two concepts, followed by a Generative Adversarial Network (GAN) to synthesize the image. The approach heavily relies on the construction of prompts,
through \emph{prompt engineering} to describe the image expected as output. 
In other words, this approach blends concepts online \emph{indirectly} through a verbal description of how they should look like.
In
contrast,~\cite{melzi2023does} approached the problem by investigating whether
conceptual blending corresponds to interpolation within the latent space of the
prompts. From their results, the authors concluded that the latent space
partially encodes conceptual blending without further training.
More recently, \citeauthor{LeKu-ISD24} \citeyear{LeKu-ISD24} extended this idea to further notions of 
\emph{in-betweenness}.

Building on this foundation, we conducted a preliminary exploration in~\cite{olearo2024blend}, which serves as
a direct precursor to this work. In that study, we introduced novel
blending strategies designed to exploit both the iterative nature of the
synthesis process and the architectural properties of the neural noise
estimator in diffusion models. This paper extends our prior work by expanding the analysis of the
blending methods, conducting a bigger and more comprehensive user study with new
and stronger metrics to evaluate the results, exploring and in-depth discussing
additional applications and implications of the proposed methods.

Independently from us and in
parallel to our research, \citeauthor{tp2o}~\citeyear{tp2o} proposed
\emph{balance swap-sampling}, a novel approach for generating blended images
from two textual concepts. Akin to~\citeauthor{melzi2023does}, this latter work exploits
the structure of the prompt latent space by embedding two concepts in a latent
space and then swapping their column vectors. Similarly, \citeauthor{swap-autoencoder}~\citeyear{swap-autoencoder} introduced so-called
\emph{swapping autoencoders} by encoding images in two separate latent spaces representing their
structure and texture. By swapping the texture latent space between two
images, the decoder is able to generate a new image with the structure of one image and
the texture of another. This approach is similar to the one proposed in our
previous work~\cite[see Section~\ref{sec:unet-pipeline}]{olearo2024blend}, but requires
training a new dedicated decoder, which is not necessary in our approach.
A zero-shot framework which aims at transferring the visual
appearance of an object to another object by leveraging its semantic knowledge was proposed by
\citeauthor{style-cross-image}~\citeyear{style-cross-image}.
This appearance transfer is achieved by exploiting the cross-attention layers~\cite{attention} of the
denoising network during the synthesis process.
Through a different line of work, \citeauthor{xiong2024novel}~\citeyear{xiong2024novel} experimented with the
similar task of \emph{text-image} fusion by balancing text and image features in
the cross-attention layer. 

Very recently, \citeauthor{freeblend} \citeyear{freeblend} introduced FreeBlend, a feedback-driven
latent interpolation method for concept blending in diffusion models.
In contrast to our approach, which represents concepts using textual prompts,
FreeBlend leverages image embeddings as input through unCLIP \cite{dalle-2}.
Their method involves injecting different embeddings at various stages of
the U-Net noise estimation process, similar to us, but incorporating an additional interpolation module and a refinement stage to
enhance fine details in the generated samples. However, as their work was published after the completion of our study, a direct comparison falls
     outside the scope of this paper.

%% file: sections/preliminaries.tex
Recent advances in generative AI have redefined the boundaries of creative computing, allowing machines to synthesize novel and diverse outputs across various domains. In particular, diffusion models, exemplified by Stable Diffusion~\cite{stable-diffusion-xl}, or DALL-E~\cite{dalle-2}, have showed remarkable capacity for generating richly detailed images via iterative denoising processes, given textual prompts. Trained on large-scale image–text datasets, these models learn latent embeddings that capture both broad visual semantics and fine-grained stylistic attributes.
In what follows, Section~\ref{subsec:diffusion_models} details the fundamentals of diffusion models, describing how iterative sampling leverages latent representations to yield high-quality images. 
We then connect these fundamentals to the concept of blending before showing (in Section~\ref{sec:blending-methods}) how a single diffusion pipeline can seamlessly merge multiple prompts into inventive, zero-shot hybrids, with no additional training.

\subsection{Diffusion Models}\label{subsec:diffusion_models}

Diffusion models \cite{dm-2015,dm-2020} have rapidly gained prominence as probabilistic generative architectures for producing high-fidelity images. Conceptually, they learn to transform noise into coherent samples through two complementary phases: (i) a forward process that gradually adds noise to an image over multiple steps, and
(ii) a reverse (denoising) process that trains a neural network to iteratively remove noise until converging on the data distribution.

Such models are well-suited for text-conditioned image (or text-to-image) generation. By injecting semantic cues, usually derived from a text encoder, into the denoising steps, diffusion pipelines can output images that reflect the content and style specified by the user's prompt. Among various text-to-image diffusion approaches, Stable Diffusion~\cite{stable-diffusion} introduces a latent-space framework to improve efficiency: images are first encoded into a lower-dimensional latent representation, and the diffusion process occurs there rather than in full-resolution pixel space. A final decoder then maps latent samples back to the visible domain.

Concretely, a common Stable Diffusion pipeline comprises: i) a Variational Autoencoder (VAE)~\cite{vaes_og} that encodes raw images into a latent space \(\mathbb{R}^n\) and decodes them back to pixels, ii) a U-Net~\cite{unet} with an encoder–decoder architecture, responsible for iteratively denoising latent representations at each timestep, with cross-attention layers that incorporate textual prompts, thus aligning the generated imagery with user-defined concepts, and iii) a text encoder, typically a CLIP model~\cite{clip}, which transforms textual prompts into prompt embeddings. These embeddings modulate the U-Net's denoising trajectory by guiding how noise is removed.

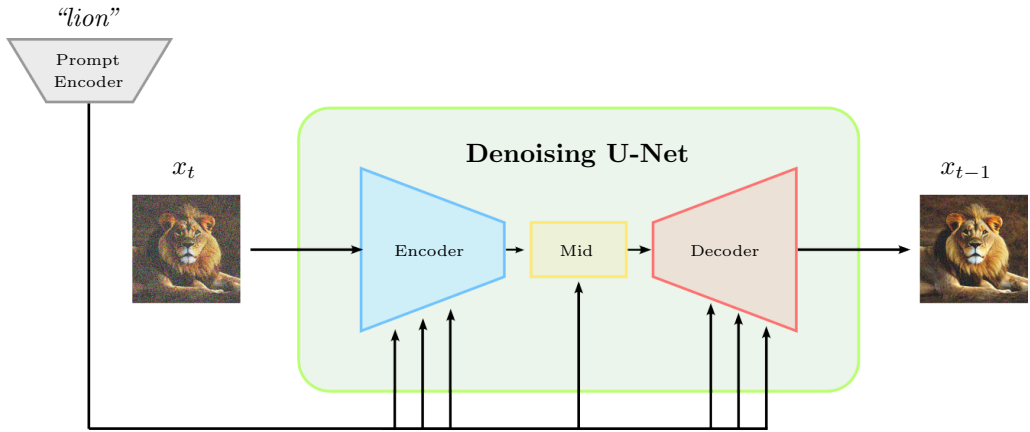
\begin{figure}[tbh!]
	\vspace{12pt}
	\centering
	\begin{overpic}[percent, grid=false, tics=10, scale=.5, width=0.89\linewidth]
		{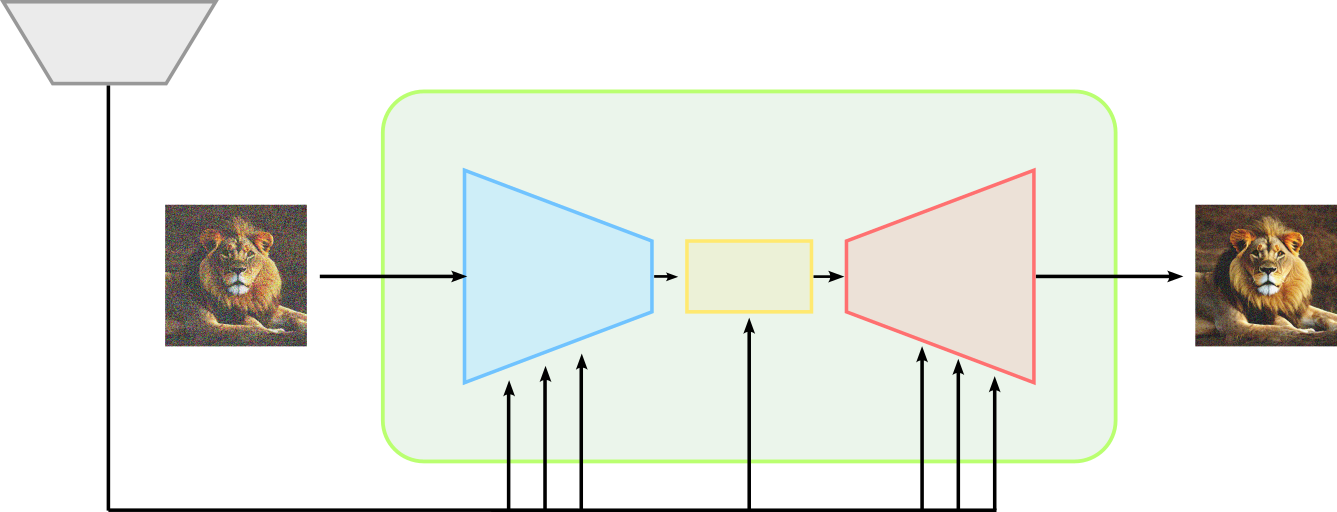}
		\put(4.4, 39.5){\small{\emph{``lion''}}}
		\put(4.8, 35.7){\tiny{Prompt}}
		\put(4.6, 33.5){\tiny{Encoder}}

		\put(38, 17){\tiny{Encoder}}
		\put(54.2, 17){\tiny{Mid}}
		\put(67, 17){\tiny{Decoder}}

		\put(45, 26.2){\small{\textbf{{Denoising U-Net}}}}

		\put(16.2, 25){\small{$x_t$}}
		\put(91.5, 25){\small{$x_{t-1}$}}
	\end{overpic}
	\caption{
		Simplified overview of the Stable Diffusion pipeline. A latent representation of an image is progressively denoised in the reverse process, with text-encoder embeddings injected at various U-Net layers to guide generation. Images eventually decode back to pixel space via the VAE.
	}
	\label{fig:pipeline-base}
\end{figure}

Figure~\ref{fig:pipeline-base} presents a simplified depiction of this pipeline. The latent representation of an image is gradually corrupted in the forward process; then, during sampling, the \mbox{U-Net} removes noise step by step, leveraging the text encoder's guidance. Once sufficiently denoised, the latent is decoded back to pixel space. By operating in a latent domain, Stable Diffusion not only reduces computation but also grants flexible prompt-based control over the generation.

In this work, we adopt Stable Diffusion v$1.4$ for its open-source availability, community validation, and established hyper-parameters. While more recent variants exist, v$1.4$ provides a reliable and transparent foundation for investigating how textual concepts can be fused or manipulated without retraining, aligning directly with our exploration of conceptual blending in a generative context.

\paragraph{Diffusion Models and Visual Concept Blending.}
Our work seeks to operationalize the conceptual blending process using text-to-image diffusion models, which learn to generate images from text prompts. 
Rather than selecting visual features or training specialized networks, we leverage zero-shot methods, manipulating prompt embeddings, latent space, or selecting components of the diffusion architecture, to merge multiple concepts into a single coherent output, producing a ``blended image'' that captures essential traits from the input concepts.

In the following sections, we describe several diffusion-based blending approaches in detail (Sections~\ref{sec:blending-methods}~and~\ref{sec:analysis}) and evaluate them through an extensive user study (Section~\ref{sec:validation-and-results}). Our investigation, thus, bridges cognitive theories of conceptual blending with large-scale generative modeling, highlighting how diffusion pipelines might be used for creative blending and where their limitations become apparent.

%% file: sections/blending-methods.tex
This section introduces four strategies for performing concept blending within diffusion models, i) \PRO, ii) \SWI, iii) \ALT, and iv) \UNE, all of which function inside the inference pipeline to avoid any additional fine-tuning. By combining prompts directly in the model’s sampling process, each method preserves recognizable features from both original concepts and enables zero-shot compositional blends, without requiring any training.
We selected these four strategies as they each exemplify a distinct level or stage at which two prompts can be merged: directly in the text-encoder embedding (\PRO), at a specific timestep mid-denoising (\SWI), on alternating timesteps (\ALT), or across the encoder–decoder split within the U-Net (\UNE). This variety sheds light on how different points of intervention in the diffusion process can shape the resulting blended image.

\begin{figure}[tbh!]
	\centering
	\begin{overpic}
		[width=0.89\linewidth]{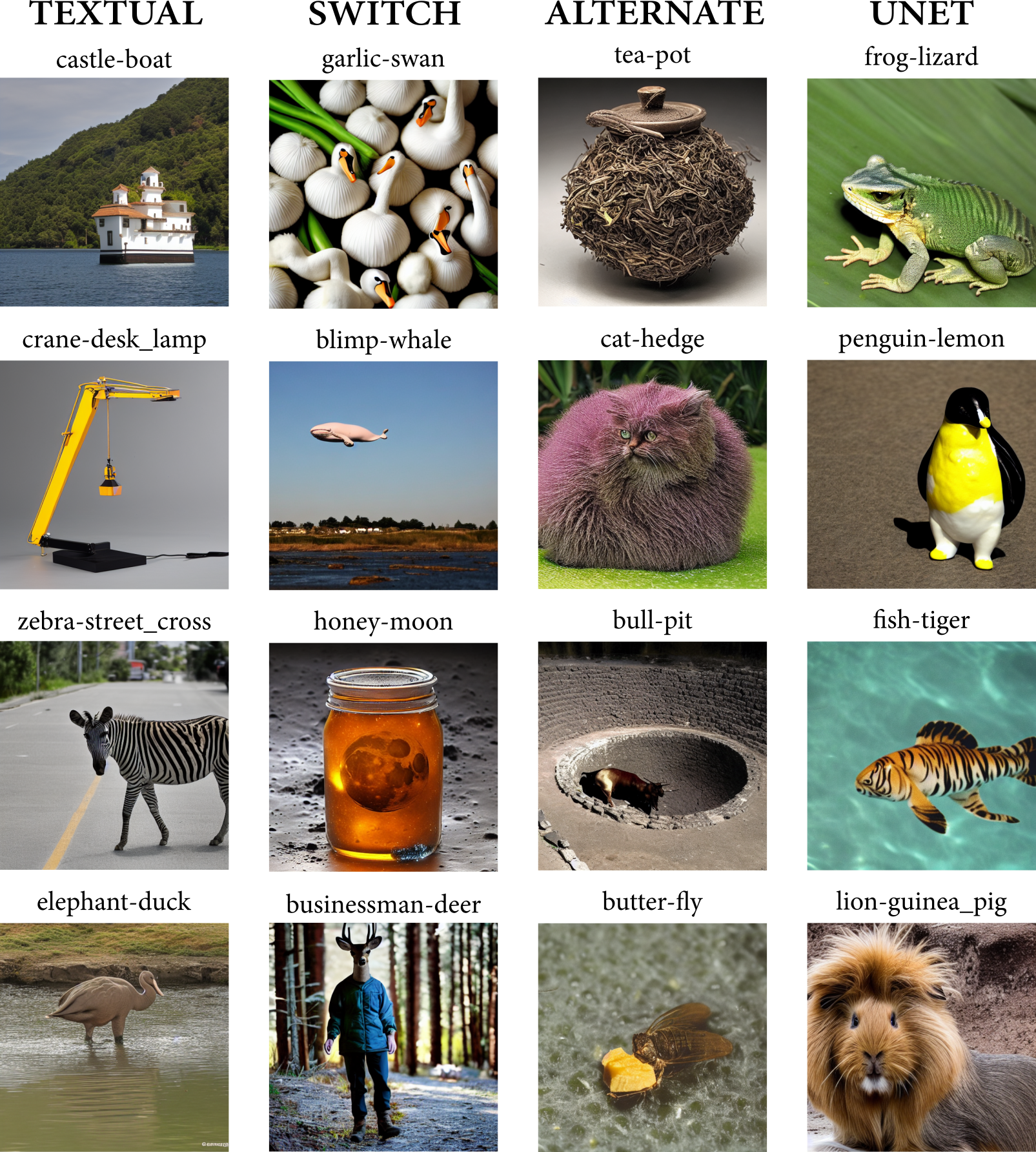}
	\end{overpic}
	\caption[Showcase of the blending methods]{
		Examples of the four blending methods, i.e., \PRO, \SWI, \ALT, and \UNE. Each resulting image is generated from two textual prompts displayed above it.
	}\label{fig:cover}
\end{figure}

Figure~\ref{fig:cover} provides a qualitative example of how each strategy synthesizes a single image from two distinct textual prompts. In brief:
\begin{itemize}
    \item \PRO merges the concepts in the text encoder’s latent space by combining the respective embeddings (e.g., via interpolation) into a single vector.
    \item \SWI begins denoising with prompt \(p_1\) and switches to \(p_2\) at a chosen timestep.  
    \item \ALT alternates the conditioning prompt at each denoising step, interweaving features from each concept across even and odd timesteps.  
    \item \UNE splits the conditioning across different U-Net stages, typically guiding the encoder/bottleneck with \(p_1\) and the decoder with \(p_2\).
\end{itemize}

The detailed methodology of each blending approach is presented in the following subsections, where we provide a step-by-step explanation of how prompts are injected or combined throughout the denoising process.

\paragraph{Experimental Setup.} 
In our experimental evaluation, we employ Stable Diffusion 
v$1.4$ \cite{stable-diffusion} with the
UniPCMultistepScheduler~\cite{zhao2023unipc} set at $25$ steps. This version
uses a fixed pretrained text encoder (CLIP Vit-L/14~\cite{clip}).
All the images are generated as $512 \times 512$ pixels with the diffusion
process carried in FP16 precision in a latent space downscaled by a factor of
$8$. The conditioning signal is provided only in the form of textual prompts and
the guidance scale is set to $7.5$ across all the experiments. 
We selected v$1.4$ over newer variants (e.g., v$1.5$, v$2.1$) as it produces high-quality outputs for our short, single-word prompts, while balancing runtime efficiency with image fidelity.

We have integrated all four blending methods under a common codebase, ensuring consistent usage of the model \(\mathcal{G}\), seeds, and prompt formatting. The repository, complete with sample outputs, prompt configurations, and environment setup details, is publicly available on GitHub.\footnote{\url{https://github.com/LorenzoOlearo/blending-diffusion-models}}  

\subsection{\PRO: Interpolation in the Prompt Latent Space}\label{sec:textual-pipeline}

\vspace{12pt}
\begin{figure}[thb]
\centering
\begin{overpic}[percent, grid=false, tics=10, scale=.5, width=0.89\linewidth]
{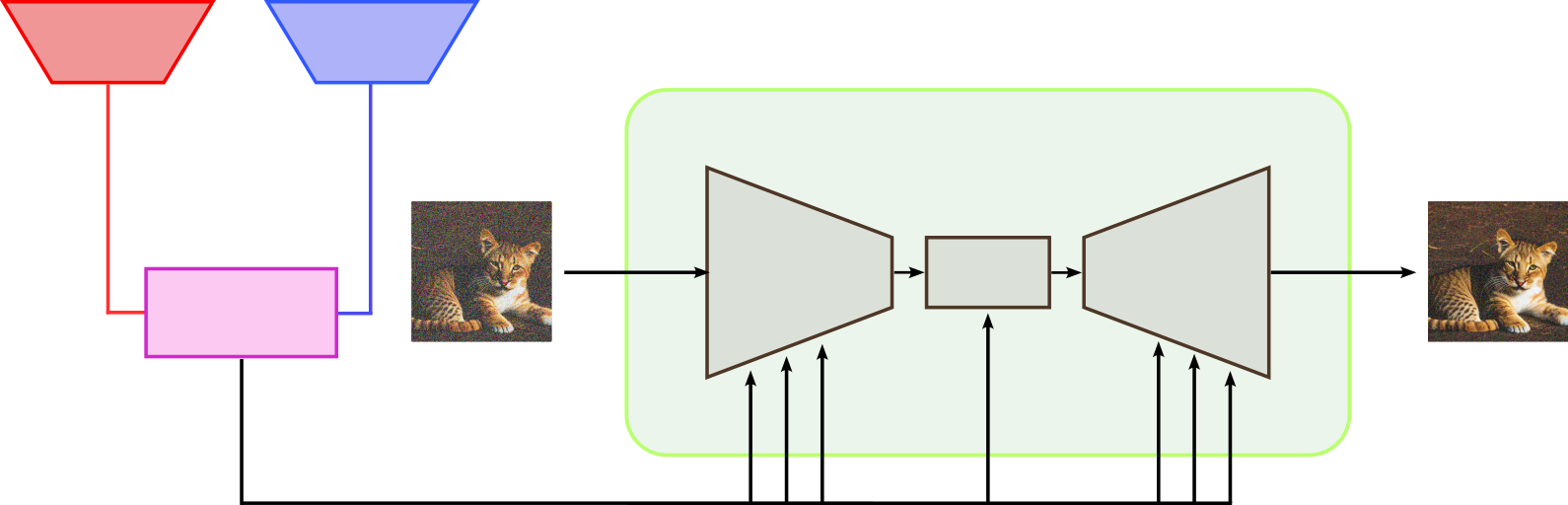}
	\put(3.8, 33.5){\small{\textit{``lion''}}}
	\put(21, 33.5){\small{\textit{``cat''}}}

	\put(4, 30.2){\tiny{Prompt}}
	\put(3.8, 28.4){\tiny{Encoder}}

	\put(21, 30.2){\tiny{Prompt}}
	\put(20.8, 28.4){\tiny{Encoder}}

	\put(8, 17){\tiny{$n p_1^* + (1-n)p_2^*$}}

	\put(28.5, 21){\small{$x_t$}}
	\put(93.5, 21){\small{$x_{t-1}$}}

	\put(53, 22.2){\small{\textbf{{Denoising U-Net}}}}

	\put(47.3, 14.3){\tiny{Encoder}}
	\put(61.5, 14.3){\tiny{Mid}}
	\put(71.9, 14.3){\tiny{Decoder}}

	\put(9.6, 11.8){\footnotesize \textit{interpolate}}
\end{overpic}
\caption{
	The \PRO~blending pipeline. The two input prompts $p_1$ and $p_2$ are
	blended by linearly interpolating their respective latent representations
	$p^{*}_1$ and $p^{*}_2$. The sample shown in this figure is generated using
	the \PRO{} pipeline, conditioned on the prompts \emph{``lion''}
	and \emph{``cat''}.
}
\label{fig:textual-pipeline}
\end{figure}
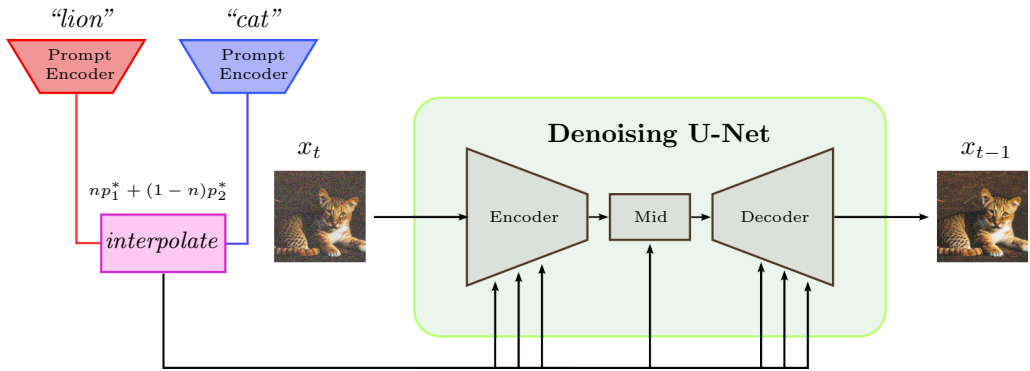

The \PRO blending method merges two prompt embeddings in the text-encoder latent space without any additional fine-tuning. Proposed by \citeauthor{melzi2023does}~\cite{melzi2023does}, it reflects the idea that vector operations on the embeddings can produce compositional effects. Formally, given two prompts \(p_1\) and \(p_2\), their latent representations \(p_1^*\) and \(p_2^*\) are first obtained from the text encoder. We then define a weighted interpolation:

\[
   p_{\mathrm{blend}}^* \;=\; \alpha \, p_1^* \;+\; (1 - \alpha)\, p_2^*,
   \quad
   \alpha \in [0, 1],
\]

where \(\alpha=0.5\) yields a simple midpoint (i.e., the Euclidean mean) between \(p_1^*\) and \(p_2^*\). This blended vector, \(p_{\mathrm{blend}}^*\), is subsequently fed into the diffusion model's cross-attention modules in the same manner as a standard single-prompt embedding would be (see Figure~\ref{fig:textual-pipeline}).

By adjusting \(\alpha\), one can shift the balance of features in the blended concept. For instance, when combining ``\emph{cat}'' \(\bigl(p_1^*\bigr)\) and ``\emph{lion}'' \(\bigl(p_2^*\bigr)\), choosing \(\alpha \approx 0.3\) emphasizes the cat-like traits more heavily, while \(\alpha \approx 0.7\) yields a lion-focused image. Throughout our experiments, we keep \(\alpha\) fixed at 0.5, to produce a balanced, symmetrical blend that does not favour either concept.

As with all blending methods in this paper, we use the same random seed across different prompts and blends, ensuring that variations in the output stem primarily from the blending itself rather than the random initialization. We note that the \PRO interpolation in latent space does not strictly guarantee a clear visual hybrid, instead, it captures a point in the semantic domain that might (or might not) correspond to a truly blended image, an inherent uncertainty reflecting the learned representation's complexity.

\subsection{\SWI: Prompt Switching in the Iterative Denoising Process}\label{sec:switch-pipeline}

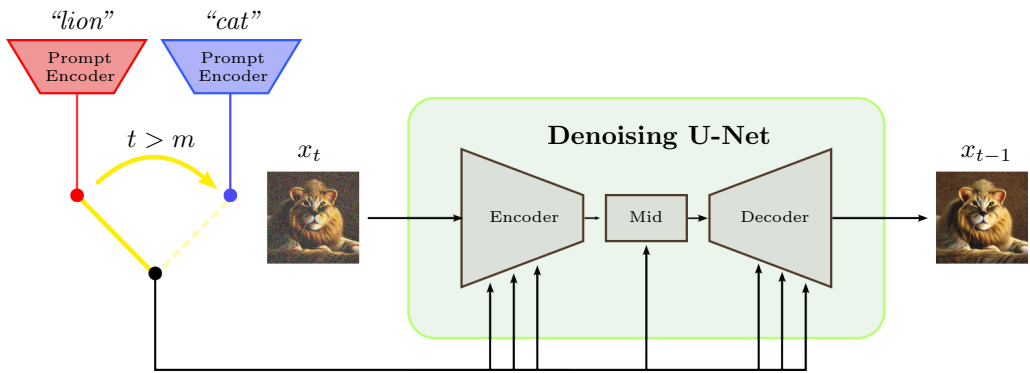
\begin{figure}[thb]
\vspace{12pt}
\centering
\begin{overpic}[percent, grid=false, tics=10, scale=.5, width=0.89\linewidth]
{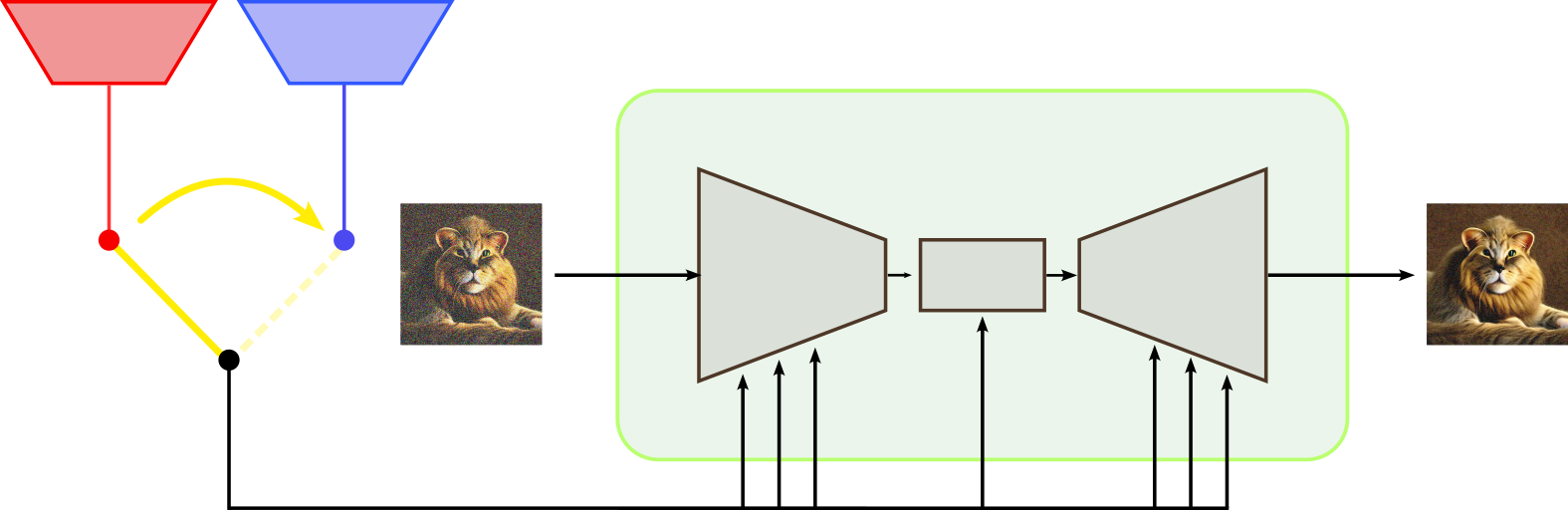}
	\put(3.8, 33.5){\small{\textit{``lion''}}}
	\put(19, 33.5){\small{\textit{``cat''}}}

	\put(4, 30.2){\tiny{Prompt}}
	\put(3.8, 28.4){\tiny{Encoder}}

	\put(19, 30.2){\tiny{Prompt}}
	\put(18.8, 28.4){\tiny{Encoder}}

	\put(11.8, 22){\small{$t > m$}}

	\put(28.5, 21){\small{$x_t$}}
	\put(93.5, 21){\small{$x_{t-1}$}}

	\put(53, 22.2){\small{\textbf{{Denoising U-Net}}}}

	\put(47.3, 14.6){\tiny{Encoder}}
	\put(61, 14.6){\tiny{Mid}}
	\put(71.9, 14.6){\tiny{Decoder}}

\end{overpic}
\caption{
    The \SWI~blending pipeline. For the initial \(m\) iterations, the U-Net is conditioned on the first prompt \(p_1\), then switched to the second prompt \(p_2\). The sample shown here uses ``\emph{lion}'' and ``\emph{cat}'' as prompts, illustrating how the global shape is derived from ``\emph{lion}'', and the finer details from ``\emph{cat}''.
}
\label{fig:switch-pipeline}
\end{figure}

The \SWI blending method first conditions the model on a prompt \(p_1\) for the initial portion of the denoising schedule, and then, at a specific timestep \(m\), switches to a second prompt \(p_2\). Conceptually, early iterations capture high-level shape and layout from \(p_1\), while later iterations integrate finer details from \(p_2\). 

Let the denoising process span \(T\) steps (defaulting to 25 in our experiments), and let \(\text{U-Net}(\mathbf{z}_t, p)\) represent the noise estimation for the sample \(x\) at timestep \(t\). The subsequent sample \(x_{t-1}\) is then computed by a scheduler, such as the UniPCMultistepScheduler~\cite{zhao2023unipc}. Then:

\begin{enumerate}
    \item Initialization: Sample a random latent \(\mathbf{z}_T\) from a noise distribution using a seed \(s\).  
    \item Denoising with \(p_1\): For timesteps \(T\) to \(m\), condition the U-Net on \(p_1\).  
    \item Prompt Switch: At timestep \(m-1\), replace \(p_1\) with \(p_2\).  
    \item Denoising with \(p_2\): Continue denoising from timestep \(m+1\) to \(t=0\), now conditioning on \(p_2\) instead.
\end{enumerate}

Formally:
\[
\mathbf{z}_{t-1} \;=\; \text{U-Net}\bigl(\mathbf{z}_t,\; p_1 \bigr)
\quad\text{for } t \geq m,
\quad\text{and}\quad
\mathbf{z}_{t-1} \;=\; \text{U-Net}\bigl(\mathbf{z}_t,\; p_2 \bigr)
\quad\text{for } t < m.
\]

A key challenge is determining the optimal switch point \(m\). Our experiments test a few values, such as 6 and 18 out of 25 (see Section~\ref{subsec:blend_ratio}). We find that if \(p_1\) and \(p_2\) produce visually dissimilar base images, switching too early can overwrite \(p_1\) almost entirely, while switching too late may fail to incorporate enough of \(p_2\). In some cases, partial blends arise if the global structure diverges significantly between the two prompts' intermediate images. 

We perform the prompt switch at the beginning of the \((m+1)\)-th denoising iteration. As a result, iterations \(1\) through \(m\) use \(p_1\), while iterations \((m+1)\) through \(T\) use \(p_2\). This prevents any overlap or partial-step mixing of prompts within a single timestep, ensuring a clean transition in the sampling routine. As with the rest of our experiments, we fix the random seed \(s\) to ensure fair comparisons across different switch timesteps and blending methods, so any variations in output can be attributed primarily to the prompt-switching mechanism itself.

Although \SWI can yield compelling hybrids (e.g., combining a ``\emph{lion}'' shape with ``\emph{cat}'' textures), abrupt changes in prompt conditioning sometimes lead to unstable transitions. Figure~\ref{fig:switch-pipeline} shows how switch merges coarse features from \(p_1\) with subtle details from \(p_2\). Still, the best setting of \(m\) can be highly prompt-dependent, highlighting the model's sensitivity to concept distance and underlying training biases.

\subsection{\ALT: Prompt Alternating in the Iterative Denoising Process}\label{sec:alternate-pipeline}

\vspace{12pt}
\begin{figure}[tbh!]
\centering
\begin{overpic}[percent, grid=false, tics=10, scale=.5, width=0.89\linewidth]
{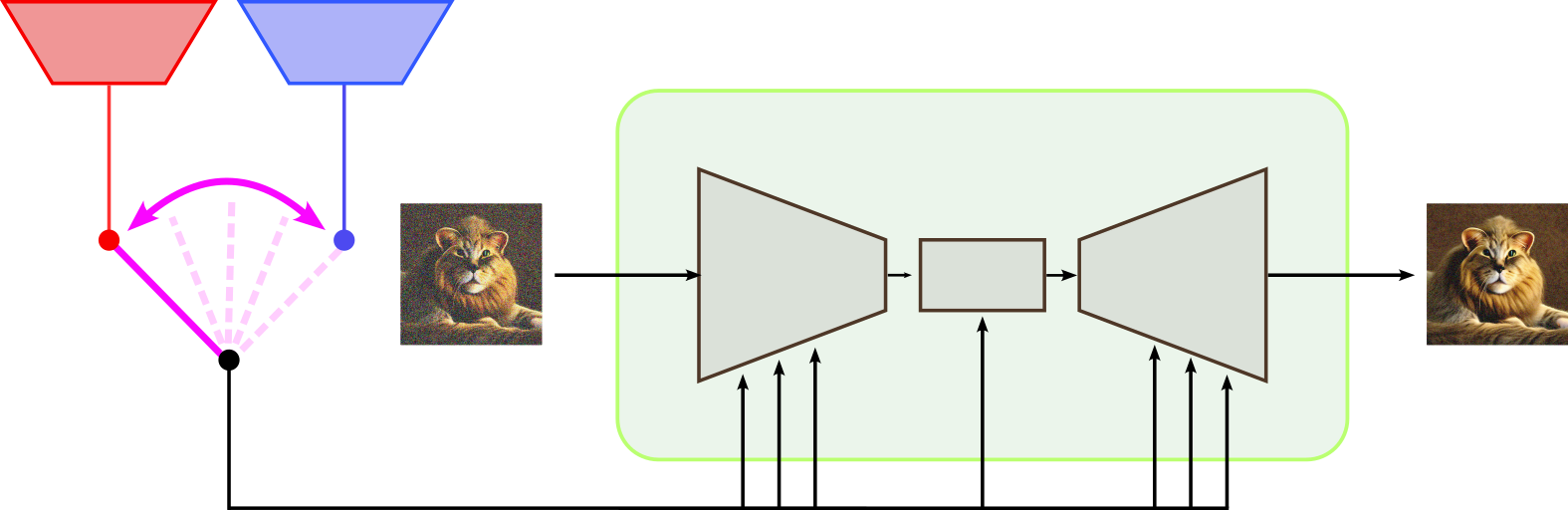}

	\put(3.8, 33.5){\small{\textit{``lion''}}}
	\put(19, 33.5){\small{\textit{``cat''}}}

	\put(4, 30.2){\tiny{Prompt}}
	\put(3.8, 28.4){\tiny{Encoder}}

	\put(19, 30.2){\tiny{Prompt}}
	\put(18.8, 28.4){\tiny{Encoder}}

	\put(9, 22){\tiny{$t \equiv 0 \mod m$}}

	\put(28.5, 21){\small{$x_t$}}
	\put(93.5, 21){\small{$x_{t-1}$}}

	\put(53, 22.2){\small{\textbf{{Denoising U-Net}}}}

	\put(47.3, 14.6){\tiny{Encoder}}
	\put(61, 14.6){\tiny{Mid}}
	\put(71.9, 14.6){\tiny{Decoder}}

\end{overpic}
\caption{
	The \ALT~blending pipeline. At the current timestep \( t \), if \( t \equiv 0 \ (\text{mod} \ m) \), the U-Net is conditioned on \( p_1 \); otherwise, it is conditioned on \( p_2 \).  
}
\label{fig:alternate-pipeline}
\end{figure}
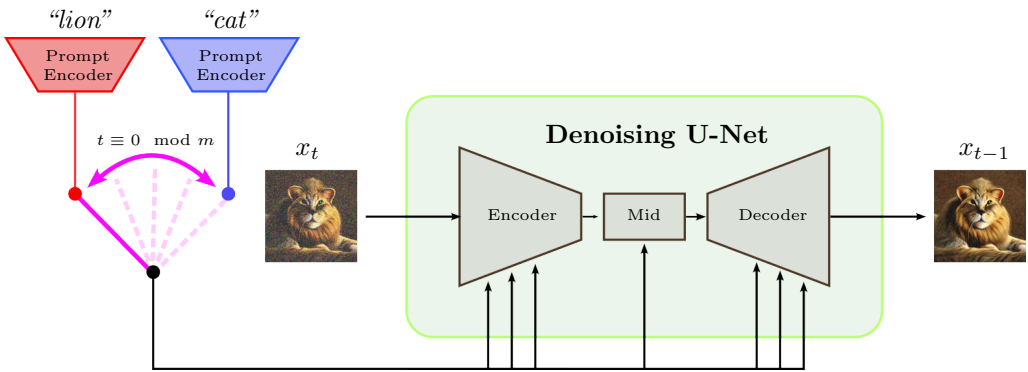

In diffusion-based architectures, the U-Net predicts the noise parametrization at each timestep \(t\), conditioning on the timestep and the prompt. The~\ALT{} blending method modifies this mechanism by presenting two different prompts in alternating order every $m$~timesteps. As illustrated in Figure~\ref{fig:alternate-pipeline}, this approach weaves features from both prompts into the final image, often producing recognizable traits of each concept.

Formally, let \(\text{U-Net}\bigl(\mathbf{z}_t, p\bigr)\) denote the denoising operation at timestep \(t\) given prompt~\(p\). The \ALT method with parameter $m$ can be summarized as:


\[
\mathbf{z}_{t+1} \;=\;
\begin{cases}
\text{U-Net}\ \!\bigl(\mathbf{z}_t, p_1\bigr) & \text{If}\ t \equiv 0 \mod m, \\
\text{U-Net}\ \!\bigl(\mathbf{z}_t, p_2\bigr) & \text{Else}.
\end{cases}
\]

By default, we assign \(p_1\) to the even timesteps and \(p_2\) to the odd timesteps, resulting in a 50-50 split across the total number of steps (25 in our experiments). This balanced scenario helps us isolate \ALT's baseline behaviour. In Section~\ref{subsec:blend_ratio}, we discuss how altering this ratio, e.g., 60–40 or 70–30, can shift the emphasis between the two concepts. As with other methods, the denoising process relies on the same random seed, ensuring that variations in output primarily reflect the degree of prompt alternation rather than chance.

In cases where \(p_1\) and \(p_2\) share a reasonably similar structure (e.g., ``\emph{lion}'' and ``\emph{cat}''), \ALT often produces blended images that highlight features from both prompts. 

A qualitative and quantitative evaluation of \ALT appears in Sections~\ref{subsec:blend_ratio}~and~\ref{sec:validation-and-results}, where we compare multiple blending ratios and prompt pairs. There, we also discuss user feedback on how effectively \ALT merges two concepts and analyse the circumstances under which it excels or fails to produce a satisfactory blend.

\subsection{\UNE: Leveraging the U-Net Conditioning}\label{sec:unet-pipeline}

\vspace{12pt}
\begin{figure}[tbh!]
\centering
\begin{overpic}[percent, grid=false, tics=10, scale=.5, width=0.89\linewidth]
{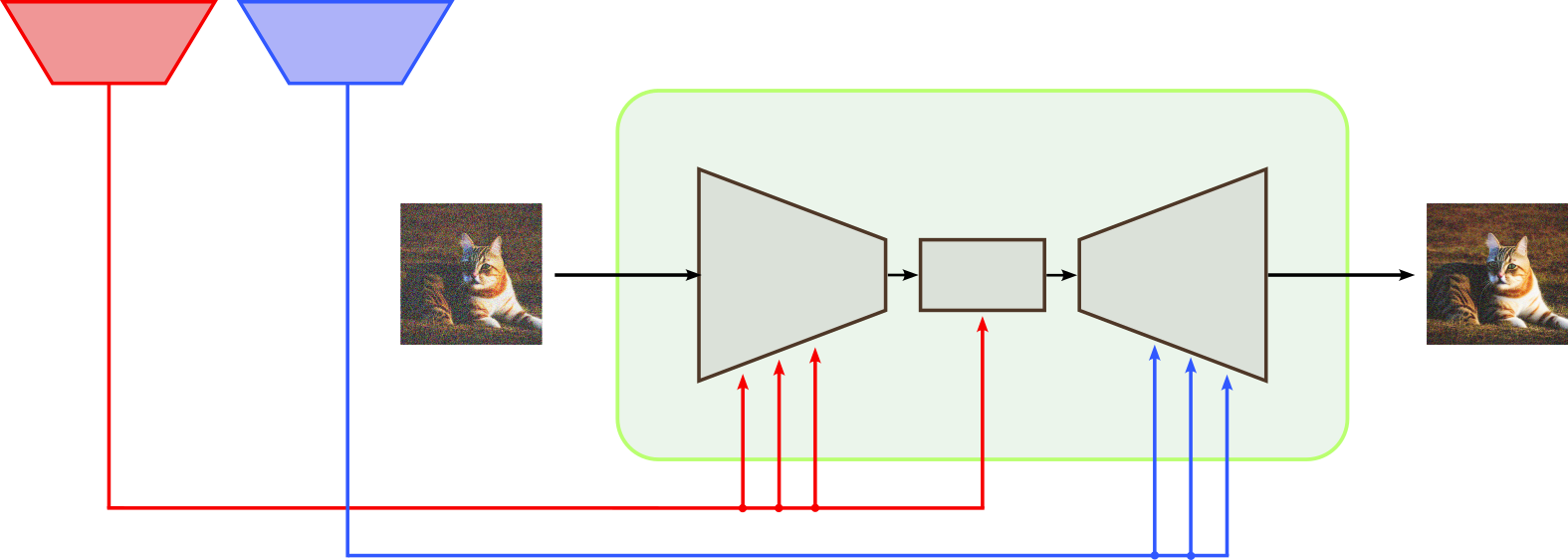}

	\put(3.8, 37){\small{\textit{``lion''}}}
	\put(19, 37){\small{\textit{``cat''}}}

	\put(4, 33.5){\tiny{Prompt}}
	\put(3.8, 31.7){\tiny{Encoder}}

	\put(19, 33.5){\tiny{Prompt}}
	\put(18.8, 31.7){\tiny{Encoder}}

	\put(28.5, 24){\small{$x_t$}}
	\put(93.5, 24){\small{$x_{t-1}$}}

	\put(53, 25.2){\small{\textbf{{Denoising U-Net}}}}

	\put(47.3, 17.6){\tiny{Encoder}}
	\put(61, 17.6){\tiny{Mid}}
	\put(71.9, 17.6){\tiny{Decoder}}
\end{overpic}
\caption{\UNE pipeline illustration. Encoder+bottleneck blocks (red) receive \(p_1^*\), while decoder blocks (blue) receive \(p_2^*\). This often retains the shape of the first concept and overlays stylistic details from the second.
}
\label{fig:unet-pipeline}
\end{figure}
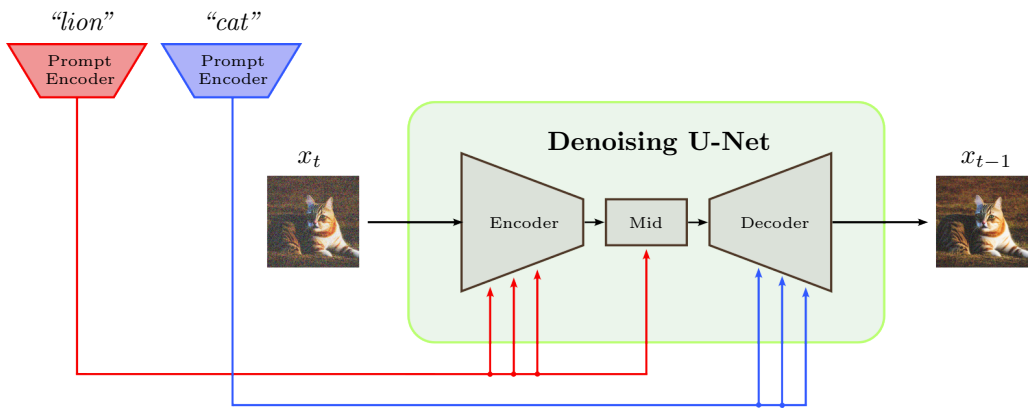

The \UNE blending method provides a novel paradigm for merging two textual prompts within the U-Net's architecture itself, rather than simply scheduling prompts or embeddings outside the U-Net (as in \PRO, \SWI, and \ALT). In those earlier strategies, either the text-encoder embedding or the timestep conditioning changed, but the underlying U-Net remained unaltered, conditioned on the same latent prompt. Here, however, we exploit the internal cross-attention blocks directly, injecting different prompts at different stages of the network.
From a conceptual blending standpoint, the encoder and bottleneck blocks serve to compress an image latent toward the target concept's features (akin to forming a ``base structure''), while the decoder reconstructs the refined latent into an output (applying ``surface'' or stylistic traits). By supplying one prompt to the encoder-bottleneck and another to the decoder, we effectively create a ``split'' that can yield images combining the form of the first concept with the style or details of the second.

In Stable Diffusion v$1.4$~\cite{stable-diffusion}, there are three cross-attention blocks in the encoder, one in the bottleneck, and three in the decoder. A single prompt embedding \(p^*\) would normally pass through all these blocks. Our method selectively switches from \(p_1^*\) to \(p_2^*\) partway through the network. This means the U-Net's internal representation can draw on concept \(p_1\) early on, and concept \(p_2\) later, yielding a blended outcome.

Unlike the prior methods, which operate ``outside'' the U-Net (e.g., interpolating embeddings or swapping prompts across timesteps), \UNE requires manipulating each cross-attention block in the model. Hence, we conducted a qualitatively study to determine where to switch from \(p_1^*\) to \(p_2^*\). As shown in Figure 7, we systematically test every possible split point among the seven cross-attention blocks (labelled \(n - m\)), qualitatively evaluating which point yields the best hybrid. This analysis shows that certain switch points (especially around the bottleneck or late encoder) maintain the global shape of \(p_1\) while injecting stylized details of \(p_2\).

\begin{sloppypar}
Formally, let \(\mathbf{z}_0\) be an initial latent and let the cross-attention blocks be \(\{E_0, E_1, E_2, B, D_0, D_1, D_2\}\). Define \(\text{split}\) as a partition indicating which blocks receive \(p_1^*\) and which receive \(p_2^*\). Then:
\end{sloppypar}

\begin{algorithm}[!h]
\begin{algorithmic}[1]
\State \(\mathbf{z} \leftarrow \mathbf{z}_0\)
\For{\(i \in \{ E_0,\dots,D_2 \}\)}
    \If{\(i\) in \texttt{split for} \(p_1\)}
        \State \(p^* \leftarrow p_1^*\)
    \Else
        \State \(p^* \leftarrow p_2^*\)
    \EndIf
    \State \(\mathbf{z} \leftarrow \text{CrossAttentionBlock}_i(\mathbf{z},\,p^*)\)
\EndFor
\State \(\mathbf{z}_T \leftarrow \mathbf{z}\)
\end{algorithmic}
\caption{A simplified pseudo-code snippet for the \UNE{} approach, showing how each cross-attention block uses \(p_1^*\) or \(p_2^*\). A fixed random seed ensures consistency when comparing different splits.}
\label{alg:unet-approach}
\end{algorithm}

\begin{figure}[tbh!]
\centering
\begin{overpic}[percent, grid=false, tics=10, scale=.5, width=0.89\linewidth]
{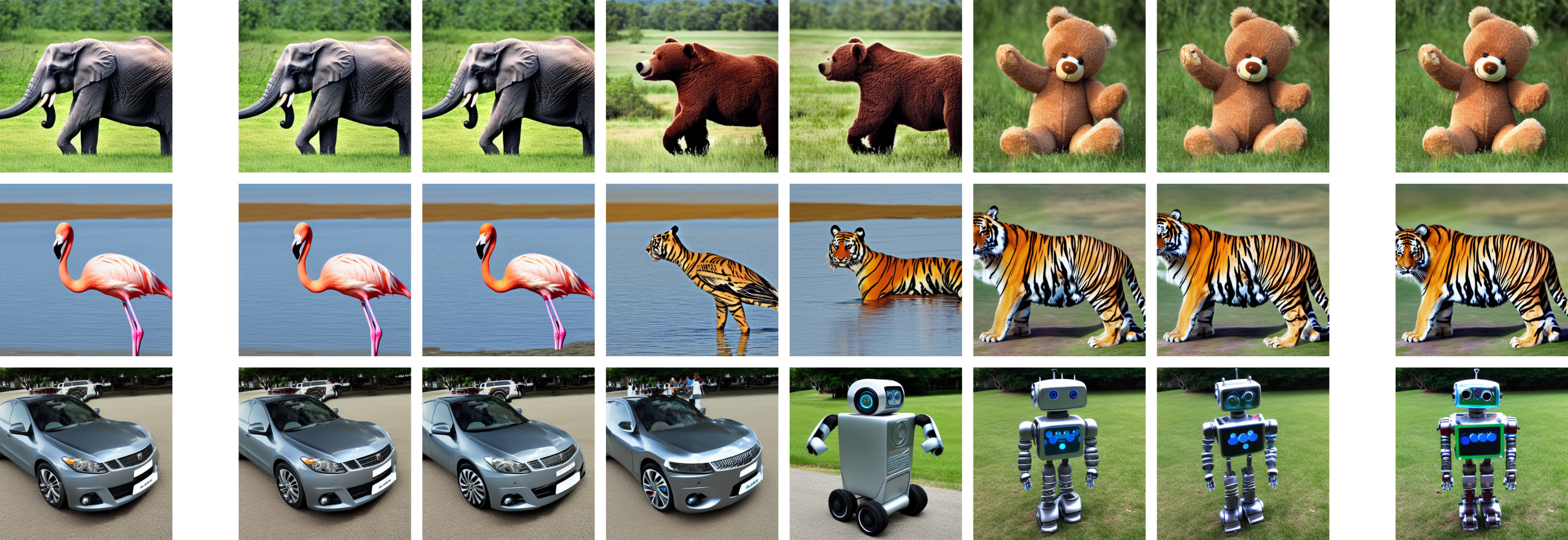}
	\put(18, 35.5){\small{$1-6$}}
	\put(29.8, 35.5){\small{$2-5$}}
	\put(42, 35.5){\small{$3-4$}}
	\put(53, 35.5){\small{$4-3$}}
	\put(64.8, 35.5){\small{$5-2$}}
	\put(76.5, 35.5){\small{$6-1$}}

	\put(-10.4, 28.5){\small{\emph{elephant}}}
	\put(-11, 17){\small{\emph{flamingo}}}
	\put(-5, 5){\small{\emph{car}}}

	\put(101, 28.5){\small{\emph{teddy bear}}}
	\put(101, 17){\small{\emph{tiger}}}
	\put(101, 5){\small{\emph{robot}}}

\end{overpic}
\caption{Qualitative results for the \UNE method. Each row uses different prompt pairs (e.g., ``\emph{elephant}'' vs. ``\emph{teddy bear}''), and columns labelled \(n - m\) show how switching from \(p_1^*\) to \(p_2^*\) at block \(n\) affects the final image. The analysis provides insights on which blocks most strongly influence shape vs. style.
}\label{fig:unet-ratio}
\end{figure}
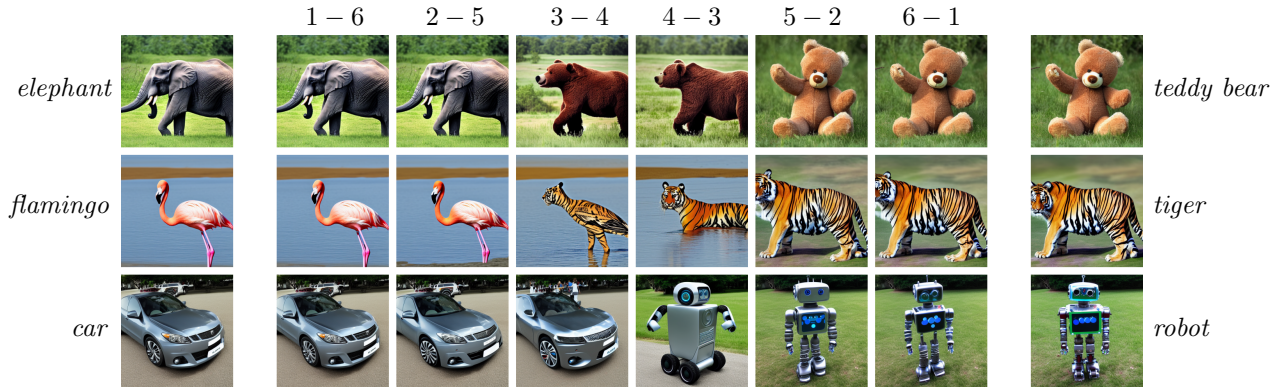

We found that encoding and bottleneck blocks guided by \(p_1^*\), followed by decoder blocks guided by \(p_2^*\), yielded stable and visually appealing hybrids for many concept pairs. Thus, we present our main results with this ``encoder and bottleneck = \(p_1\), decoder = \(p_2\)'' configuration (as depicted in Figure~\ref{fig:unet-pipeline}). However, Figure~\ref{fig:unet-ratio} illustrates that not all concept pairs behave uniformly, some pairs benefit from switching earlier or later. Furthermore, while this analysis highlights the architectural sensitivity of this approach, the best split may also depend on seed variance and the semantic distance between prompts. As we discuss in Section~\ref{sec:analysis}, these factors can cause the output to lean more heavily toward one prompt.

\vspace{12pt}

In contrast to \PRO, \SWI, or \ALT, where we alter embeddings outside the denoising network or at the timesteps, \UNE inherently depends on the internal layout of the model's cross-attention blocks. We show a systematic comparison of \UNE with the other three methods in Section~\ref{sec:analysis}, where both qualitative examples and user-study feedback help assess the relative strengths and weaknesses of this internal conditioning approach.

%% file: sections/analysis.tex
Having introduced four distinct strategies for concept blending with text-to-image diffusion models in Section~\ref{sec:blending-methods}, we now examine the key dependencies that shape how these methods perform in practice. Throughout this section, we discuss with qualitative examples and user-study observations why certain prompt pairs or parameter modifications lead to coherent blends, whereas others result in unexpected or one-sided images.
Specifically, we discuss how swapping the order of two prompts (e.g., ``\emph{lion}'' vs. ``\emph{cat}'' or ``\emph{cat}'' vs. ``\emph{lion}'') can favour one concept visually, depending on the method's internal biases or scheduling approach (Section~\ref{subsec:symmetry}), how the blend ratio between two concepts can emphasize one concept more heavily than the other (Section~\ref{subsec:blend_ratio}), how random seed variations can produce radically different results, even for the same prompt pair and blending parameters (Section~\ref{sec:seed-dependency}), and the limitations (Section~\ref{subsec:limitations}) of this work that can alter the outcome of zero-shot blending.

\subsection{Symmetry}
\label{subsec:symmetry}

In conceptual blending theory \cite{FaTu-08}, a main concept typically provides the core structure, which is then enriched by modifier concepts that introduce additional attributes. The resulting blend captures much of the main concept's essence while integrating features from the secondary one. In our setting, \PRO is the only purely symmetric method: it averages (or interpolates) two embeddings, so reversing the prompt order does not affect the outcome. By contrast, the other three methods, i.e., \SWI, \ALT, and \UNE, exhibit inherent asymmetry, meaning that swapping the order of two prompts often produces notably different images.

This asymmetry becomes particularly visible with compound words like pitbull, whose modern usage (a dog breed) differs from its historical roots (``bull in a pit''). For instance, when blending ``\emph{pit}'' and ``\emph{bull}'' via \SWI, which always begins denoising with the first prompt, or \UNE, which conditions the encoder on the first prompt and the decoder on the second, selecting which prompt is principal determines the final appearance. Even compound words such as ``rain bow'' or ``jelly fish'' can show similar order-sensitivity, where the ``main concept'' might dominate shape and colour. As illustrated in Figure~\ref{fig:same-seed}, \PRO and \ALT tend to prioritize the second prompt, while \SWI and \UNE tend to privilege the first. This unexpected divergence led us to assign the first prompt as the main concept and the second as the modifier to maintain consistency. Consequently, we generate ``\emph{bull-pit}'' rather than ``\emph{pit-bull}''.

\begin{figure}[tbh!]
	\centering
	\begin{overpic}[percent, grid=false, tics=10, scale=.5, width=0.89\linewidth]
		{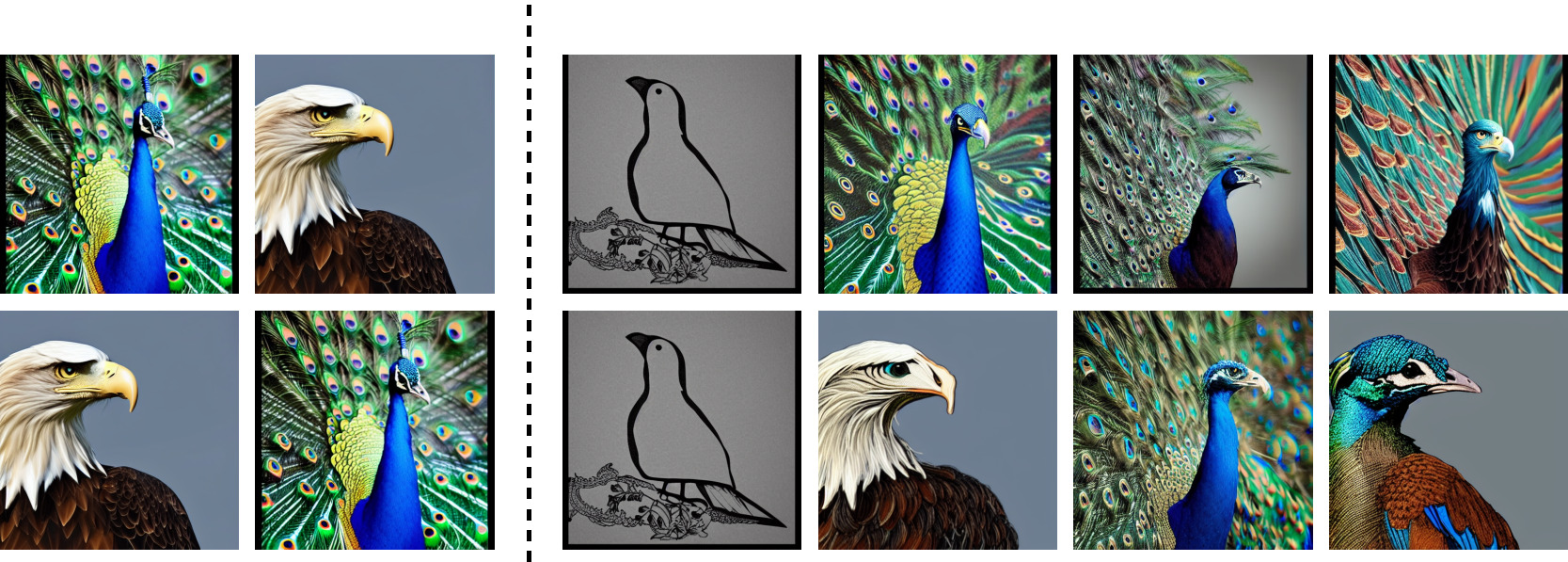}
		\put(38.7, 33.5){\small{\PRO}}
		\put(55.9, 33.5){\small{\SWI}}
		\put(70, 33.5){\small{\ALT}}
		\put(89.4, 33.5){\small{\UNE}}

		\put(-4.5, 24){\small{(a)}}
		\put(-4.5, 8){\small{(b)}}
	\end{overpic}
	\caption{Comparison between (a) blending ``\emph{peacock}'' with ``\emph{eagle}'' and (b) reversing the order to ``\emph{eagle}'' with ``\emph{peacock}''. All images share the same initial noise.
	}
	\label{fig:simmetry}
\end{figure}
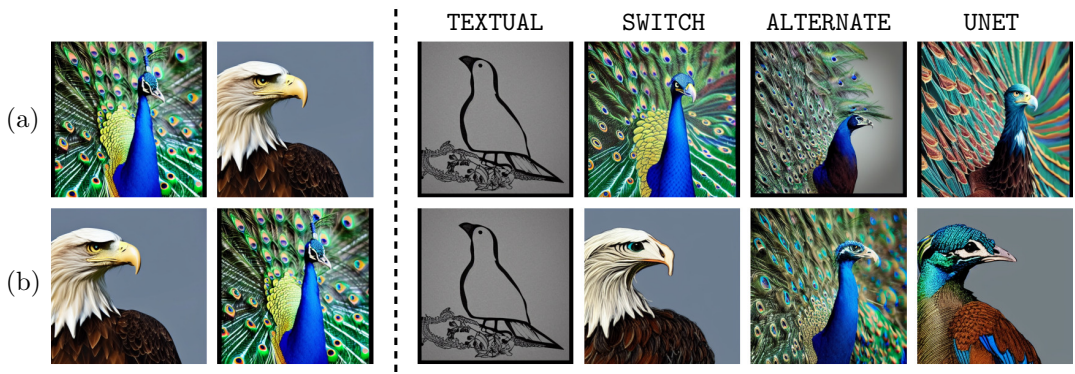

Figure~\ref{fig:simmetry} further shows this phenomenon with the concepts ``peacock'' and ``eagle''. Reversing the order can lead to a peacock adopting an eagle's beak or an eagle adopting peacock plumage. Such outcomes highlight a main–modifier dynamic shaped by each method's architecture: \PRO's vector averaging is order-invariant, \ALT partially shifts focus to the second prompt, and \SWI or \UNE systematically bias the image toward whichever prompt is introduced first. Although this may be beneficial for practical creative tasks, for example, a user intentionally favouring one concept, unaware designers could be surprised by how drastically prompt order alters results. Our user study (see Section~\ref{sec:validation-and-results}) corroborates that participants notice these asymmetries, often judging images based on which concept they perceive as the core. Moreover, the effect is generally more pronounced with single-word prompts, where reversing the main–modifier assignment fully flips the blended outcome. More complex or multi-word prompts may compound this asymmetry even further, introducing additional variability in which concept takes precedence.

\subsection{Blend Ratio}
\label{subsec:blend_ratio}

\begin{figure}[tbh!]
	\vspace{12pt}
	\centering
	\begin{overpic}[percent, grid=false, tics=10, scale=.5, width=0.89\linewidth]
		{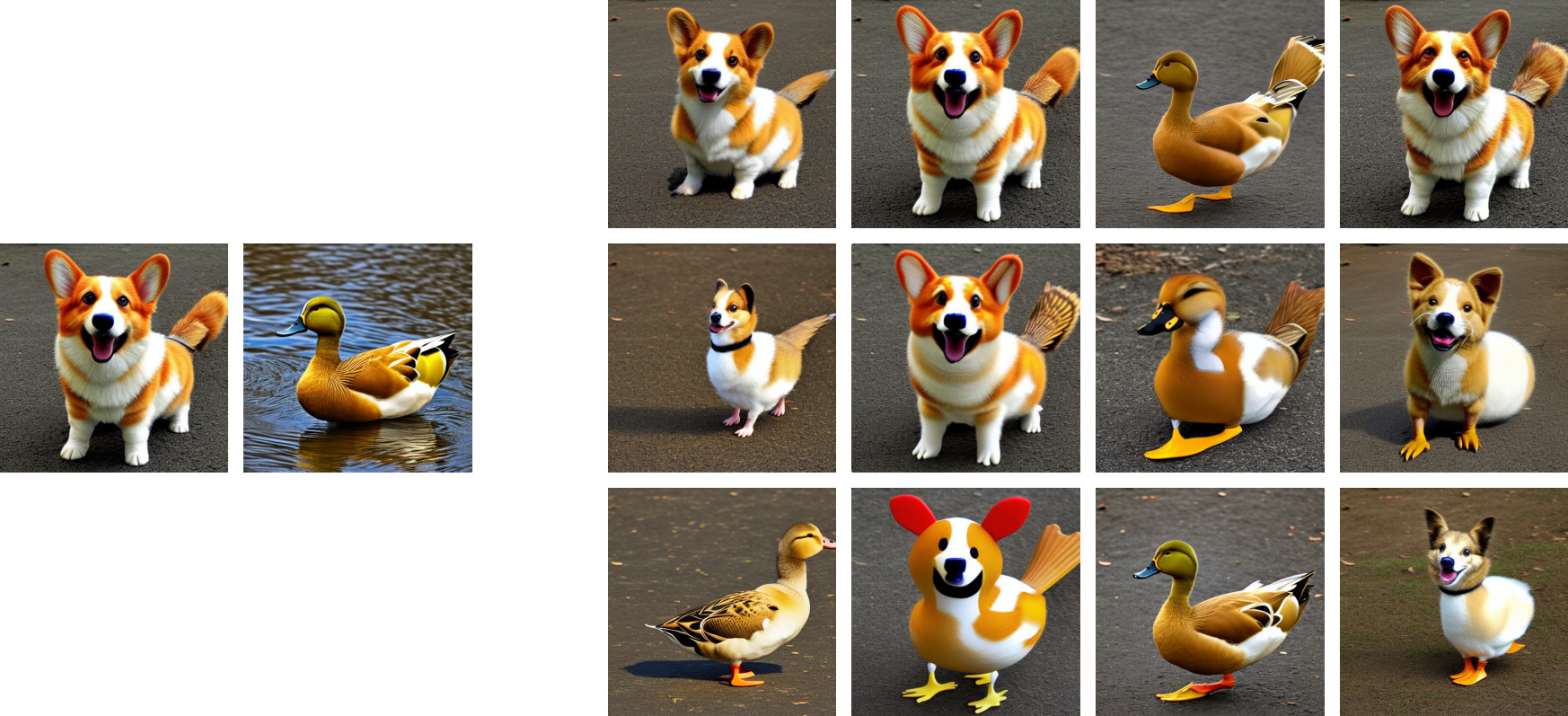}
		\put(2, 31){\small{\emph{corgi dog}}}
		\put(20, 31){\small{\emph{duck}}}
		\put(41, 46.5){\small{\PRO}}
		\put(57, 46.5){\small{\SWI}}
		\put(71, 46.5){\small{\ALT}}
		\put(90, 46.5){\small{\UNE}}
		\put(34.5, 38){\small{(a)}}
		\put(34.5, 22){\small{(b)}}
		\put(34.5, 7.5){\small{(c)}}
	\end{overpic}
	\caption{
		Comparison between the blending of the concepts \emph{``corgi dog''}
		and \emph{``duck''} with different blending ratios. Row \textbf{(a)}
		$25\%$ \emph{``corgi dog''} and $75\%$ \emph{``duck''}, row \textbf{(b)}
		both concepts are equally weighted, and row \textbf{(c)}
		$75\%$ \emph{``corgi dog''} and $25\%$ \emph{``duck''}. The same seed is
		used across all samples.
	}
	\label{fig:blend-ratio}
\end{figure}
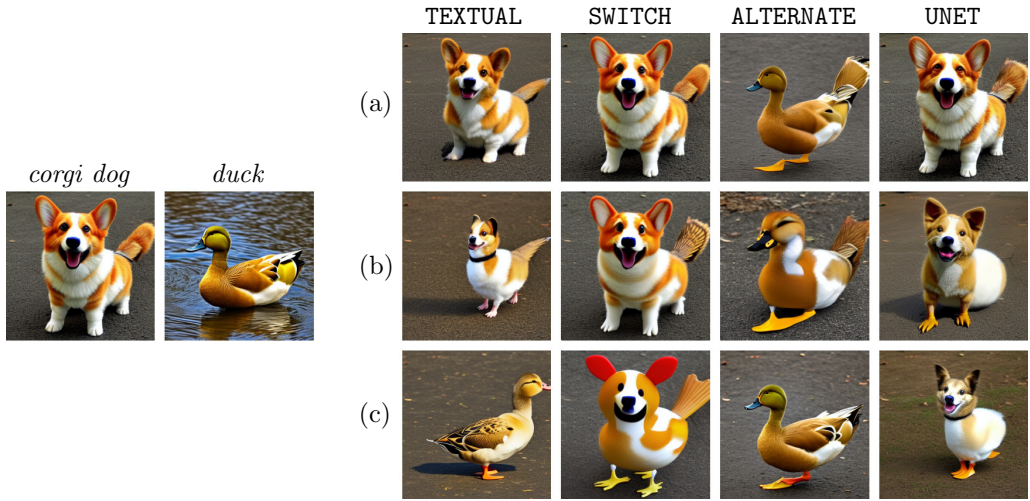

When blending two concepts into a single image, it is often desirable to control the relative emphasis each concept exerts. This blend ratio can be viewed as the proportion of conceptual features drawn from each prompt. Although each method presented in Section~\ref{sec:blending-methods} offers some mechanism to adjust this ratio, they differ in how that adjustment is realized and how the resulting effect can be interpreted. Figure~\ref{fig:blend-ratio} shows a few examples in which the ratio between ``\emph{corgi dog}'' and ``\emph{duck}'' is varied.

From a conceptual blending perspective, it can be defined as selectively adjusting how much of each concept is depicted into the blend projection, where only certain features of each concept are retained,  emphasizing the ``base'' concept while subtly weaving in attributes from the ``modifier'', or vice versa.

\paragraph{\PRO.} Of all the methods, \PRO provides the most direct ratio control, since it uses linear interpolation between prompt embeddings. Specifically, it defines a blended vector \(p_{\mathrm{blend}}^*\) as \(p_{\mathrm{blend}}^* = \alpha\, p_1^* + (1 - \alpha)\, p_2^*\), where \(\alpha \in [0,1]\). Setting \(\alpha=0.5\) weights both prompts equally, while more extreme values favour one concept almost entirely.

\paragraph{\SWI.} In the \SWI method, a single timestep \(m\) determines the switch from the first prompt to the second. Shifting \(m\) closer to the start yields a blend dominated by \(p_2\), whereas delaying the switch gives more weight to \(p_1\). This is straightforward to tune in principle but can lead to \textit{cartoonification} if the switch is abrupt. As seen in row (c) of Figure~\ref{fig:blend-ratio}, excessively late switching sometimes induces a loss of high-frequency details.

\paragraph{\ALT.} The \ALT method interleaves each prompt on alternating timesteps of the diffusion process. Controlling the ratio thus involves allocating more timesteps to \(p_1\) or \(p_2\). For instance, dedicating \(60\%\) of the timesteps to \(p_1\) and \(40\%\) to \(p_2\) yields a correspondingly skewed blend.

\paragraph{\UNE.} The \UNE approach is the most complex, as it manipulates which layers of cross-attention see each prompt. In principle, a \(25{-}75\%\) blend can be achieved by conditioning one or two early layers on \(\mathbf{p}_1^*\) and the remaining layers (including the bottleneck) on \(\mathbf{p}_2^*\). However, an odd number of cross-attention blocks requires choosing the middle block, i.e., the bottleneck, to be used for one prompt or the other.
As discussed in Section~\ref{sec:unet-pipeline}, we opt to encode the ``base'' concept in the encoder+bottleneck, then overlay features from the second prompt during the decoder. This design echoes the conceptual blending idea of a ``main structure'' with added ``modifier'' details.

Figure~\ref{fig:blend-ratio} shows how each method's ratio-control mechanism can produce distinct outcomes. \PRO yields a smooth interpolation in embedding space, \SWI enforces a sharp change at a chosen timestep, \ALT distributes timesteps between the two concepts, and \UNE changes which cross-attention blocks are conditioned on each prompt. While all methods allow ratio adjustments, their internal processes, and potential pitfalls (later discussed in Section~\ref{subsec:limitations}), require careful consideration when trying to balance two concepts precisely.

\subsection{The Seed Dependency}~\label{sec:seed-dependency}
A notable complexity of diffusion models is their reliance on the random noise from which images are sampled \cite{eml-seed}. In a standard text-to-image diffusion pipeline with classifier-free guidance \cite{free-guidance-neurips}, the initial noise strongly influences colour palettes, object poses, and other low-level traits that persist in the final image, regardless of the semantic content of the prompt. Blending methods are not exempt from this effect: even though they merge two concepts, the seed-specific features can still dominate certain aspects of the generated outcome.

While our user study (see Section~\ref{sec:validation-and-results}) does not explicitly address seed variability, we highlight here how consistent each blending method is across different seeds. In practice, we tested ten distinct random seeds for each concept pair. Methods like \UNE tend to produce subtle yet stable blends across seeds, whereas \ALT sometimes yields no discernible fusion at all but can, on rare occasions, generate highly appealing hybrids. These observations align with a broader consensus in classifier-free guided diffusion, where similar prompts produce visually similar images when initialized with the same noise, not necessarily in semantic details, but in attributes like pose, colour gradients, or compositional structure.

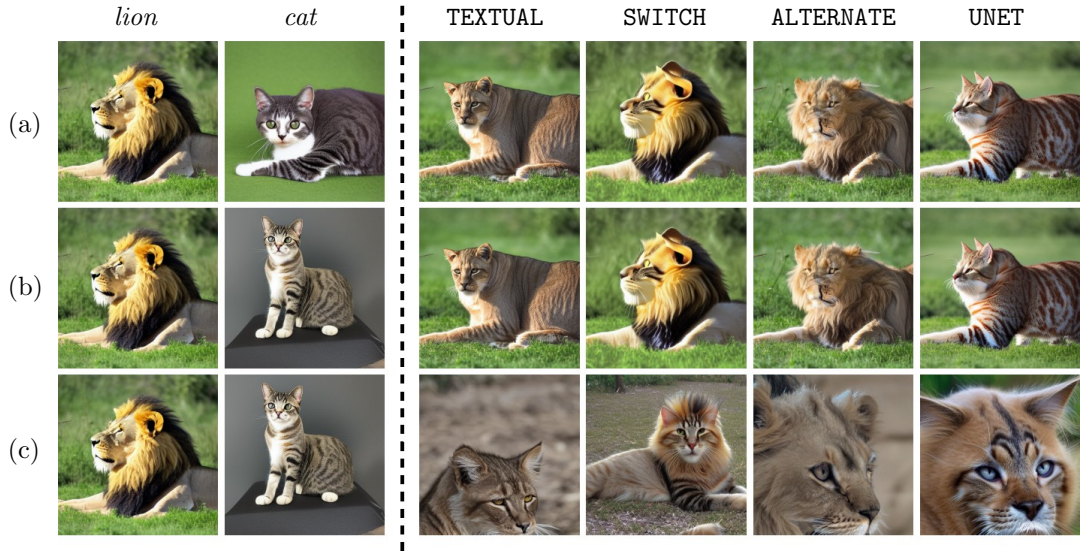
\begin{figure}[tbh!]
	\centering
	\begin{overpic}[percent, grid=false, tics=10, scale=.5, width=.89\linewidth]
		{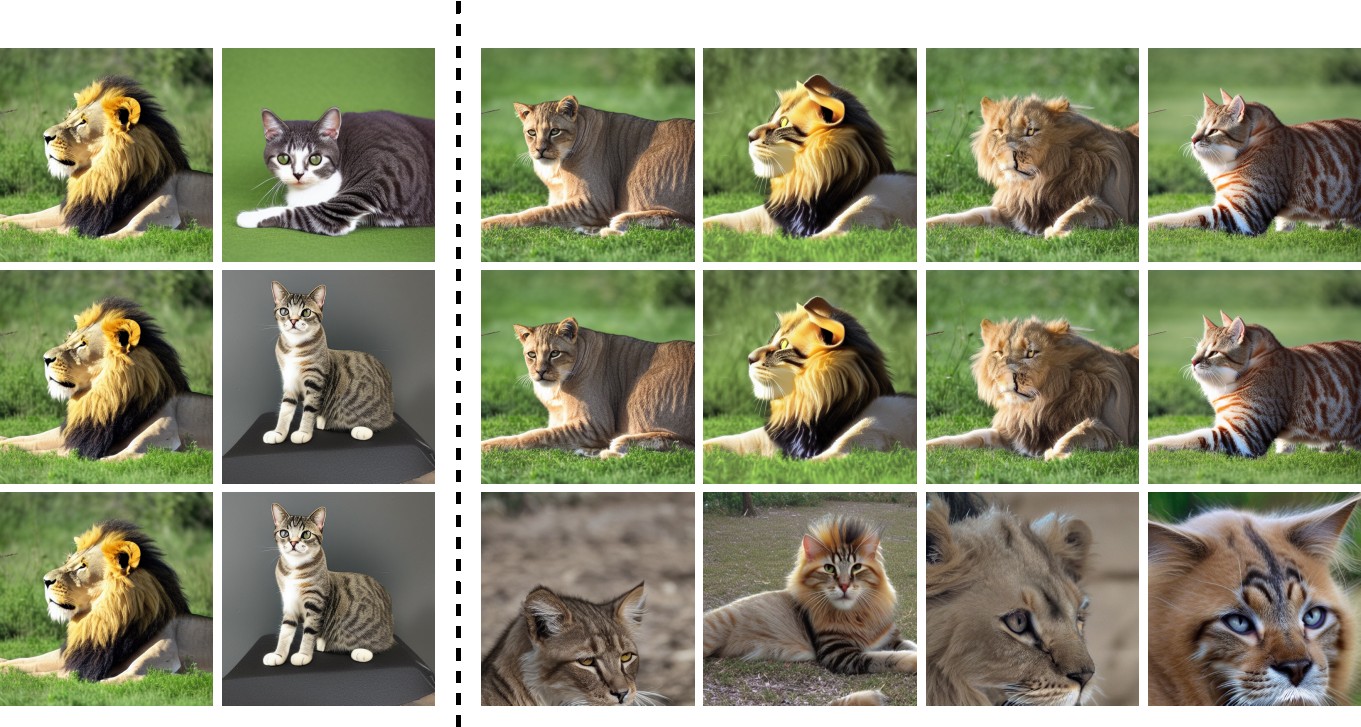}
		\put(5.4, 51.5){\small{\emph{lion}}}
		\put(22, 51.5){\small{\emph{cat}}}

		\put(38, 51.5){\small{\PRO}}
		\put(55.3, 51.5){\small{\SWI}}
		\put(69.8, 51.5){\small{\ALT}}
		\put(89.2, 51.5){\small{\UNE}}

		\put(-5, 41){\small{(a)}}
		\put(-5, 25){\small{(b)}}
		\put(-5, 9){\small{(c)}}
	\end{overpic}
	\caption{
		Comparison of results from blending the prompts ``\emph{lion}'' and
		``\emph{cat}'': \textbf{(a)} all images are generated from the same initial
		noise, \textbf{(b)} individual prompts are generated from different
		noise, but blending starts from the same noise as the first image,
		and \textbf{(c)} both individual prompts and the blending process begin
		with different noises.
	}
	\label{fig:same-vs-different}
\end{figure}

In our blending context, this implies that if we generate a ``\emph{lion}'' and a ``\emph{cat}'' image from the same seed separately, the two images often share broad visual or spatial similarities, factors that then carry over into a subsequent blend. For consistency and to focus on the semantic aspects of blending (rather than seed-induced variance), we fixed the same seed for each prompt pair. Figure~\ref{fig:same-vs-different} illustrates how the blending output changes when we vary the seeds for individual prompts, for the blend itself, or both. Figures~\ref{fig:same-seed} and~\ref{fig:full-lion-cat} show further examples of how the ``\emph{lion-cat}'' combination behaves under different seeds, highlighting how the blending methods maintain a consistent main–modifier dynamic across seeds.

\subsection{Limitations}
\label{subsec:limitations}

Each blending method discussed in this paper inevitably comes with methodological and architectural constraints that can affect the quality, consistency, and interpretability of the resulting images. Beyond these specific issues, the broader challenge lies in ensuring that conceptual blending via neural embeddings aligns well with cognitive expectations derived from classical theories such as Fauconnier and Turner’s \cite{FaTu-08}.

\PRO relies on linear interpolation in the text-encoder's geometric space, yet we have no guarantee that such interpolation mirrors how humans conceive a smooth transition between concepts. Figure~\ref{fig:blend-ratio}, for instance, shows cases where different ratios depict near–single-concept images, reflecting the mismatch between embedding-space geometry and user intuitions. \SWI can cause cartoonification when the prompt changes late in the sampling process, as illustrated in row (c) of Figure~\ref{fig:blend-ratio} and further discussed in Section~\ref{sec:validation-and-results}. Determining precisely when to flip the prompt remains a non-trivial task. \ALT occasionally yields results that lack aesthetic appeal: highly skewed split points can fragment the final image rather than integrate it, especially for simple or short prompts (e.g. ``\emph{cat}'', ``\emph{lion}''). \UNE, while powerful, is challenging to control because assigning each prompt to different cross-attention blocks can result in subtle or incomplete blends, and is sensitive to architectural specifics like the number of cross-attention layers (Section~\ref{sec:unet-pipeline} and Figure~\ref{fig:unet-ratio}).

Beyond the methods themselves, our reliance on Stable Diffusion v1.4 imposes further constraints. We selected this model for its relative simplicity and manageable computational demands, but newer pipelines or fine-tuned architectures may handle short, general prompts (the kind tested here) more reliably. Likewise, the CLIP text encoder \cite{clip}, trained on large-scale image-caption pairs, can struggle with minimal prompt descriptions.

An additional limitation emerges from the subjectivity of judging blended images. What one user sees as an elegant synergy, another may perceive as awkward. Our user-study findings (Section~\ref{sec:validation-and-results}) hint at how personal aesthetic preferences influence whether a given blend is considered ``successful''. This effect also interacts with the number and complexity of prompts tested: while we focus on short, single-word prompts (often compound-word splits) to highlight core blending behaviours, more elaborate or multi-word prompts might expose different limitations or require more complex methods of ratio and seed control.

Although we consider these constraints significant, they do not undercut the overall potential of zero-shot concept blending via text-to-image diffusion. To showcase each method's inherent strengths (rather than flaws from model or prompt limitations), we selected their best outcomes for comparative analysis. Our best outcomes were chosen through a combination of qualitative inspection by the authors and pilot user feedback, ensuring we highlight each technique's characteristic performance under favourable conditions. Ultimately, we expect future refinements, such as adopting next-generation diffusion models, more advanced text encoders, or domain-specific fine-tuning, to address many of these issues, further bridging the gap between embedding space manipulations and human-like conceptual integration.

%% file: sections/intangible.tex
In this section, we focus on abstract and subjective prompts, where one concept might represent a specific artistic style, an emotional descriptor, or a cultural theme. This reformulation expands the zero-shot blending paradigm as a style transfer method \cite{style-cross-image}, but without the additional training or specialized architectures that style-transfer approaches often require.

From a conceptual blending perspective, abstract ideas function much like intangible ``modifiers'', merging their stylistic or emotional essence with the core structure of a more concrete concept. In a text-to-image context, this is highly practical: styles are frequently used via prompt engineering, yet as \cite{style-cross-image} show, an object's appearance can be transferred to another concept in a purely zero-shot manner, leveraging diffusion models without fine-tuning. We build on this idea using the same four methods (\PRO, \SWI, \ALT, and \UNE) discussed in Section \ref{sec:blending-methods}, simply substituting intangible prompts (e.g., a painting style or an emotional expression) for one of the two concepts.

Sections 6.1, 6.2, and 6.3 explore how intangible concepts, such as paintings, emotions, or architecture, can modify a tangible subject. Although these tasks bear similarities to classical style transfer, our approach maintains the flexibility and compositional properties of zero-shot blending.

\subsection{Paintings}
\label{subsec:paintings}
One of the most compelling applications of intangible blending involves mimicking the style of a famous painting in a given scene. Notable examples include van Gogh's bold, swirling patterns, Monet's delicate impressions, and Picasso's geometric abstractions. In this setting, we use the painting (identified by its title and author) as the modifier, applying its unique look to a main concept that defines the core content of the image. This choice differs from earlier experiments (e.g., blending lion and cat), which used broader, more general concepts with many possible visual realizations. Instead, each painting has a well-defined style and composition that we expect a diffusion model to have learned during its large-scale training.

\begin{figure}[t!]
    \centering
    \vspace{12pt}
    \begin{overpic}[percent, grid=false, tics=10, scale=.5, width=1\linewidth]
        {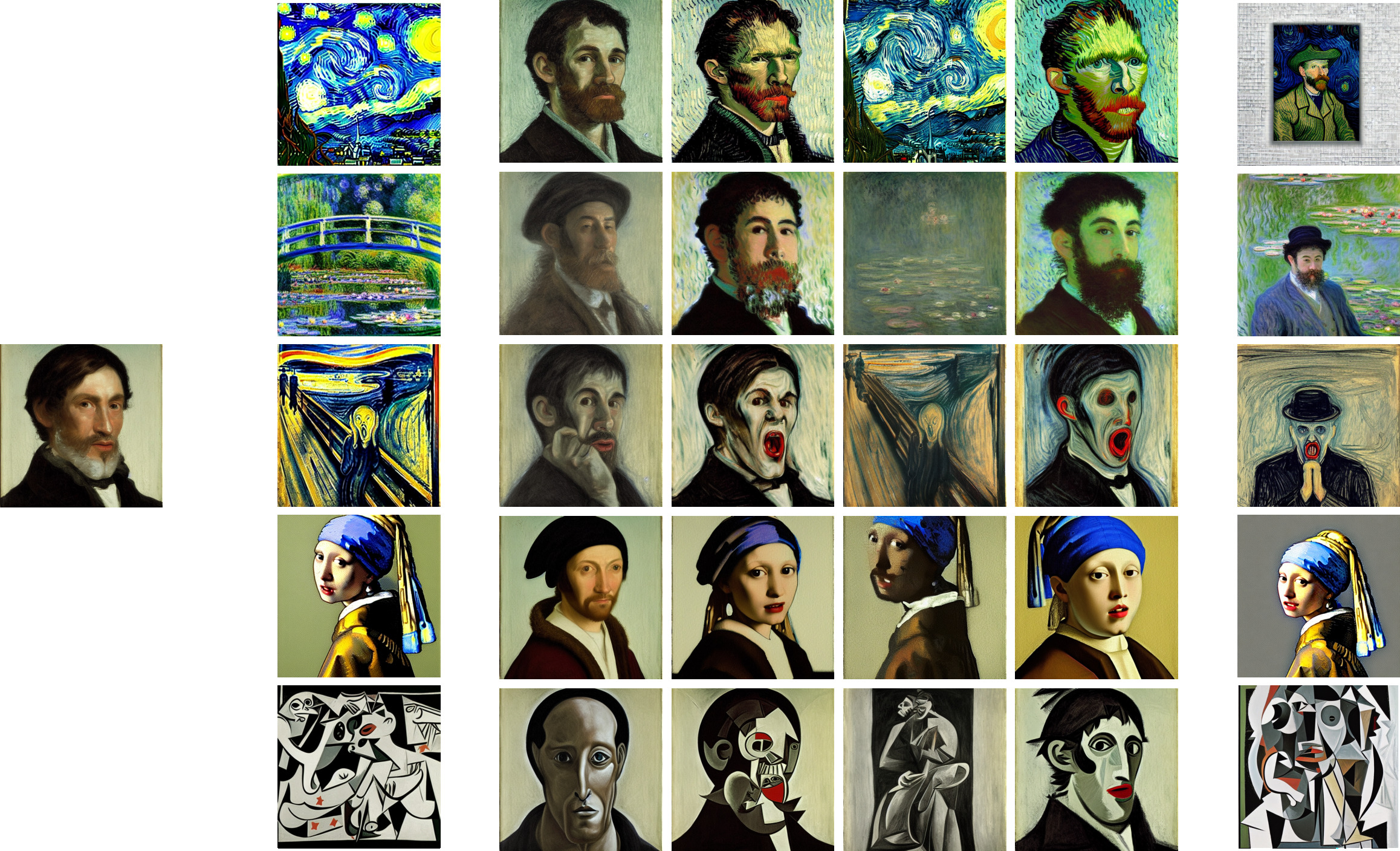}
        \put(-5, 37.5){\small{\emph{A portrait of a man}}}

        \put(37.2, 62){\small{\PRO}}
        \put(49.7, 62){\small{\SWI}}
        \put(60.5, 62){\small{\ALT}}
        \put(76, 62){\small{\UNE}}

        \put(88.8, 62){\small{\texttt{prompting}}}

        \put(15.5, 54.4){\small{(a)}}
        \put(15.5, 43){\small{(b)}}
        \put(15.5, 30.3){\small{(c)}}
        \put(15.5, 18){\small{(d)}}
        \put(15.5, 5){\small{(e)}}

    \end{overpic}
    \caption{Style blending examples for the main concept ``\emph{A portrait of a man}'', combined with (a) ``\emph{Starry Night by van Gogh}'', (b) ``\emph{Water Lilies by Claude Monet}'', (c) ``\emph{The Scream by Edvard Munch}'', (d) ``\emph{Girl with a Pearl Earring by Johannes Vermeer}'', and (e)~``\emph{Guernica by Pablo Picasso}''. The last column shows the baseline result using a single ``\emph{in the style of}'' prompt. All images are generated with the same random seed.
    }
    \label{fig:blending-style}
\end{figure}
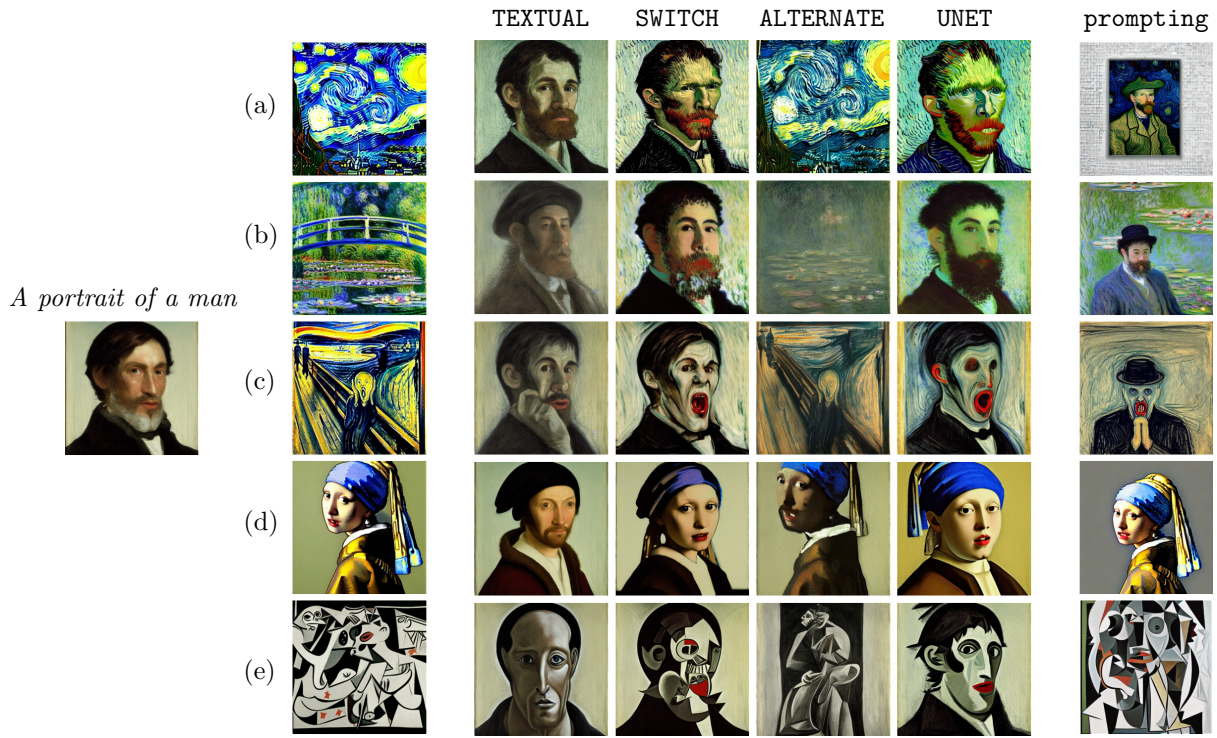

We selected five iconic paintings for their visually distinct characteristics, i) ``\emph{Starry Night by van Gogh}'', ii) ``\emph{Water Lilies by Claude Monet}'', iii) ``\emph{The Scream by Edvard Munch}'', iv) ``\emph{The Girl with a Pearl Earring by Johannes Vermeer}'', and v) ``\emph{Guernica by Pablo Picasso}''. Thus, the diffusion model's learned representation would be tested against diverse colour palettes and strong stylistic signals. In this setup, we treat each painting (described by its title and artist) as the modifier prompt, blending it with a main concept that defines the subject's semantic content, such as ``A portrait of a man''. Figure~\ref{fig:blending-style}contrasts these blended results with a baseline approach that simply appends ``\emph{in the style of [painting]}'', i.e.~if the main concept is ``\emph{A portrait of a man}'' and the desired style ``\emph{Water Lilies by Claude Monet}'', the prompt would be ``\emph{A portrait of a man in the style of Water Lilies by Claude Monet}''.

Overall, \SWI and \UNE stand out in integrating the painting's style without losing the underlying content. For instance, several methods (except \ALT) alter the man's face to resemble Van Gogh's self-portraits, hinting that the model strongly associates Van Gogh's style with a particular facial structure, likely due to extensive training images of his self-portraits. We observed no parallel effect for the other paintings, underscoring how specific artist and subject mappings can emerge from large-scale training data. 

\subsection{Emotions}
\label{subsec:emotions}
\begin{figure}[t!]
    \centering
    \vspace{12pt}
    \begin{overpic}[percent, grid=false, tics=10, scale=.5, width=1\linewidth]
        {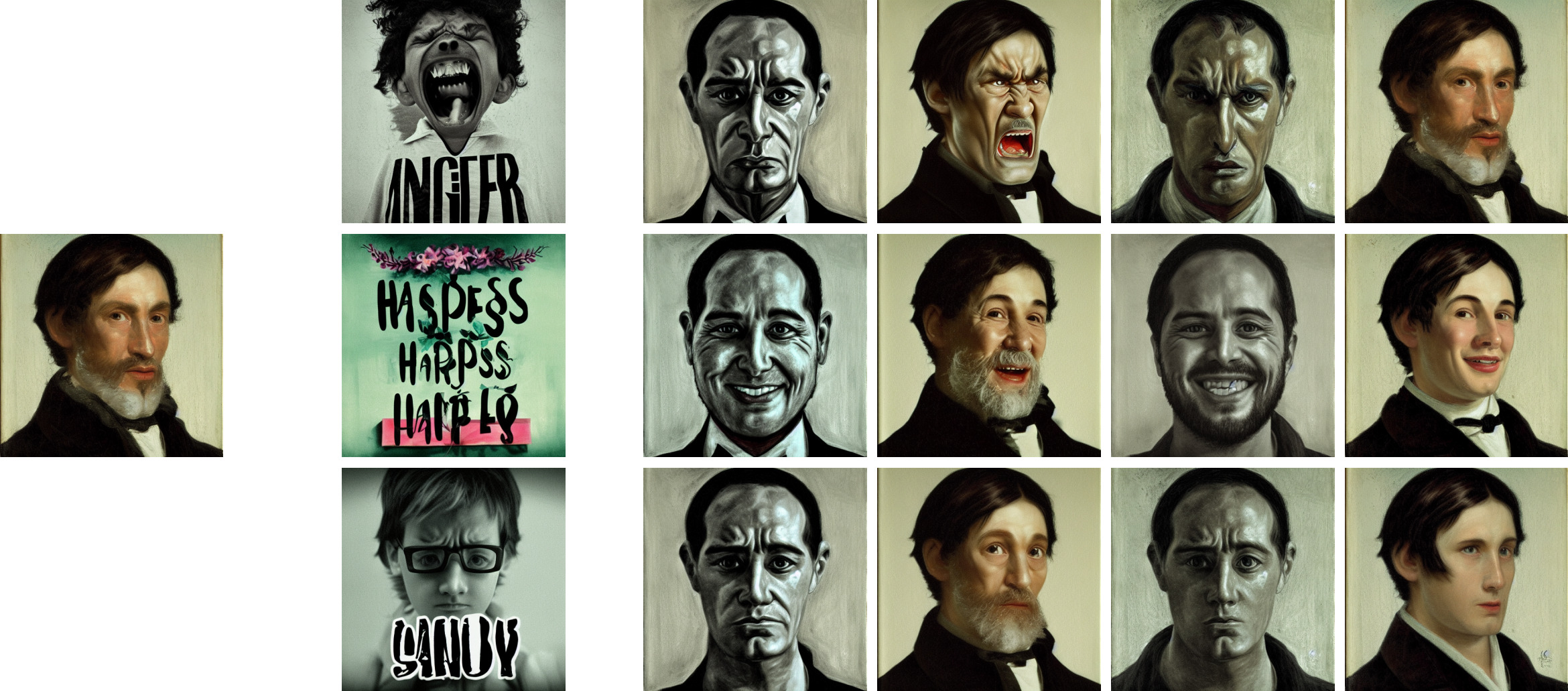}
        \put(-4, 31){\small{\emph{A portrait of a man}}}

        \put(44, 45){\small{\PRO}}
        \put(59.3, 45){\small{\SWI}}
        \put(72.3, 45){\small{\ALT}}
        \put(90.5, 45){\small{\UNE}}

        \put(17.5, 36){\small{(a)}}
        \put(17.5, 21){\small{(b)}}
        \put(17.5, 6.6){\small{(c)}}

    \end{overpic}
    \caption{
       Samples from blending the main concept ``A portrait of a man'' with three emotions: (a) ``anger'', (b) ``happiness'', and (c) ``sadness''. All images share the same initial random seed.
    }
    \label{fig:emotions}
\end{figure}
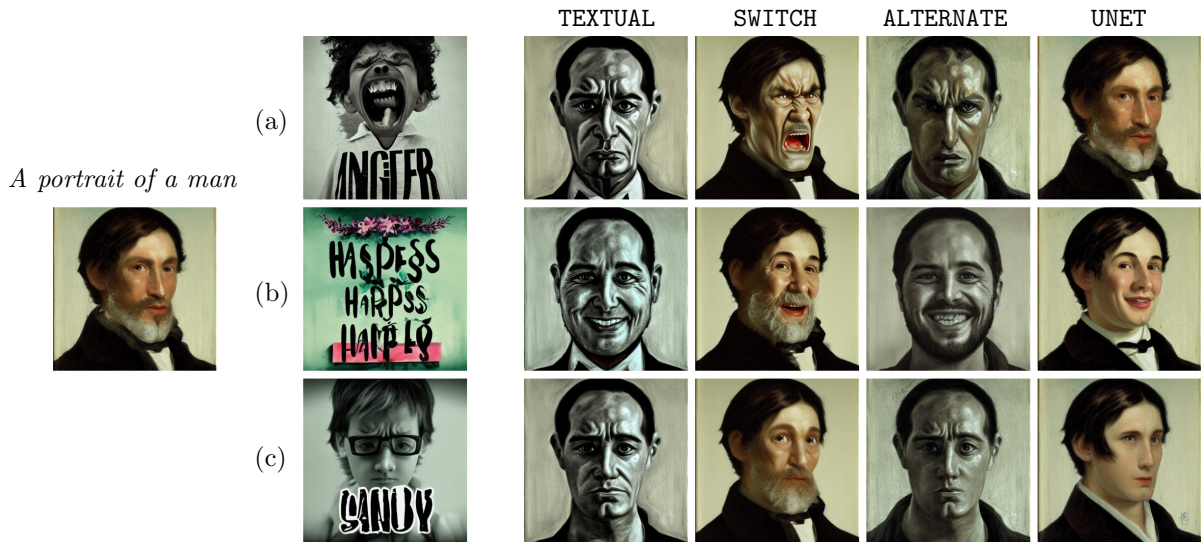
A further application of style blending is modulating the mood of an image by mixing a visual subject with an emotion. Following the style blends discussed earlier, we now set the main concept to ``\emph{A portrait of a man}'' and the modifier to a simple emotion word (e.g., ``\emph{anger}'', ``\emph{happiness}'', and ``\emph{sadness}''). 

As illustrated in Figure~\ref{fig:emotions}, each blending method successfully evokes the target emotion, although the ``best'' outcome may depend on personal preference.
Nevertheless, a few distinct behaviours emerge: \SWI and \UNE typically edit the subject's features to reflect the desired emotional state, e.g., furrowing brows or reshaping the mouth, while \PRO and \ALT may replace the subject entirely, producing a more drastic transformation. In \PRO's case, this follows from the learned latent space of the text encoder (Section~\ref{sec:textual-pipeline}); a single-word prompt like ``sadness'' can correspond to multiple visual contexts in the large-scale CLIP training set. Yet when paired with a clear prior (like ``\emph{A portrait of a man}''), it can guide the pipeline to generate a meaningful emotional display.

Overall, we highlight how zero-shot blending can capture subtle affective cues, despite the abstract nature of emotions compared to well-defined painting styles. Finally, while style-based blending and emotion-based blending share certain similarities, affective transformations are arguably less predictable, given that emotions are inherently more subjective than established artistic styles. The next subsection (Section~\ref{subsec:architecture}) extends intangible blending to architecture, examining how building aesthetics can act as a conceptual modifier for otherwise conventional prompts.

\subsection{Architecture}
\label{subsec:architecture}

\begin{figure}[tbh!]
    \vspace{12pt}
    \centering
    \begin{overpic}[percent, grid=false, tics=10, scale=.5, width=0.89\linewidth]
        {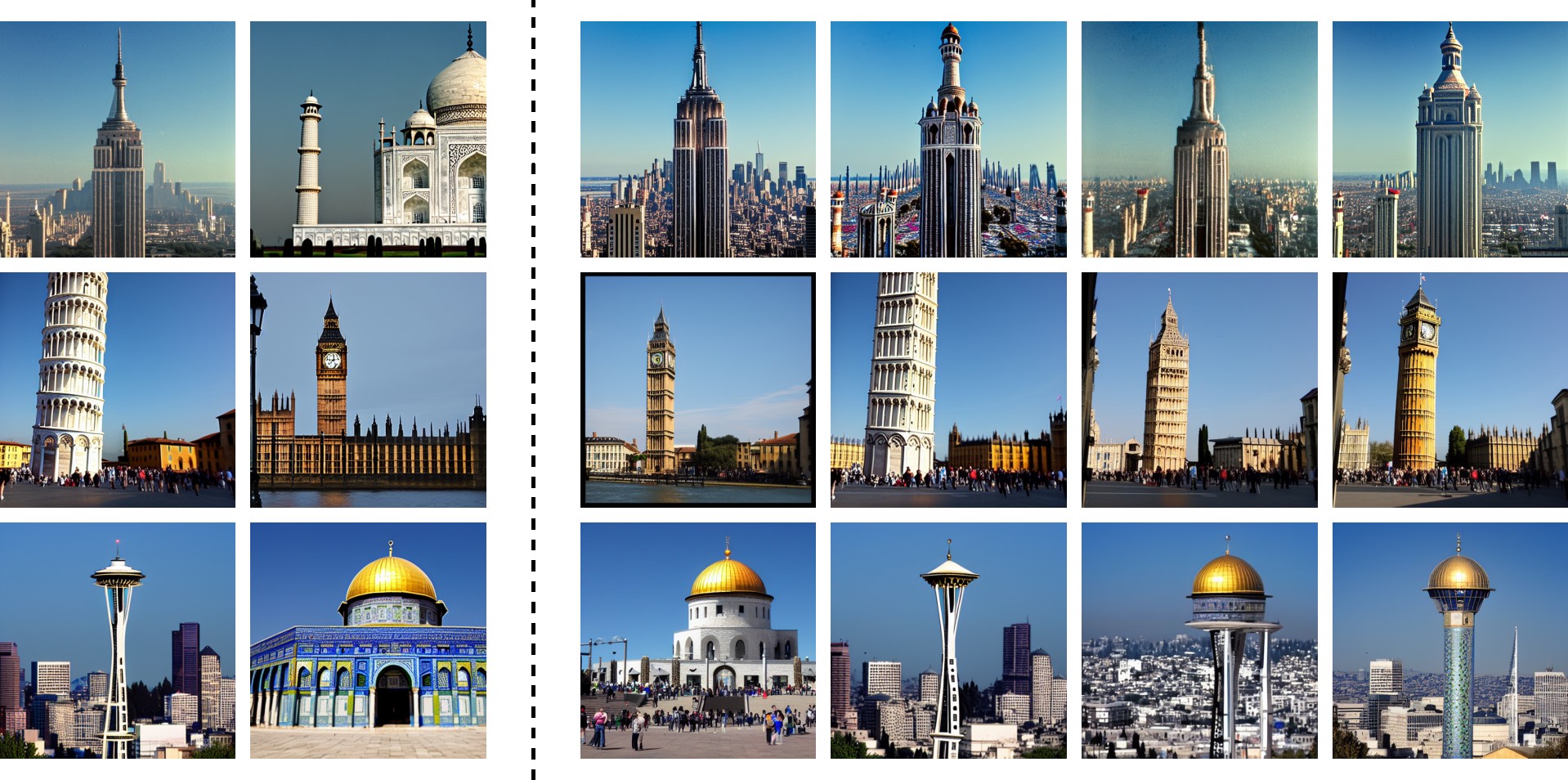}
        \put(40, 49.5){\small{\PRO}}
        \put(56.2, 49.5){\small{\SWI}}
        \put(70.5, 49.5){\small{\ALT}}
        \put(89.8, 49.5){\small{\UNE}}
        \put(-5, 39.6){\small{(a)}}
        \put(-5, 24){\small{(b)}}
        \put(-5, 7.8){\small{(c)}}
\end{overpic}
    \caption{
        Results from blending two architectural landmarks. Row (a) shows the
        blending of ``\emph{Empire State Building}'' with ``\emph{Taj Mahal}'',
        (b) ``\emph{Leaning Tower of Pisa}'' with ``\emph{Big Ben}'', and (c)
        ``\emph{Seattle Space Needle}'' with ``\emph{Dome of the Rock}''. All images share the same initial random seed.
    }
    \label{fig:architecture}
\end{figure}
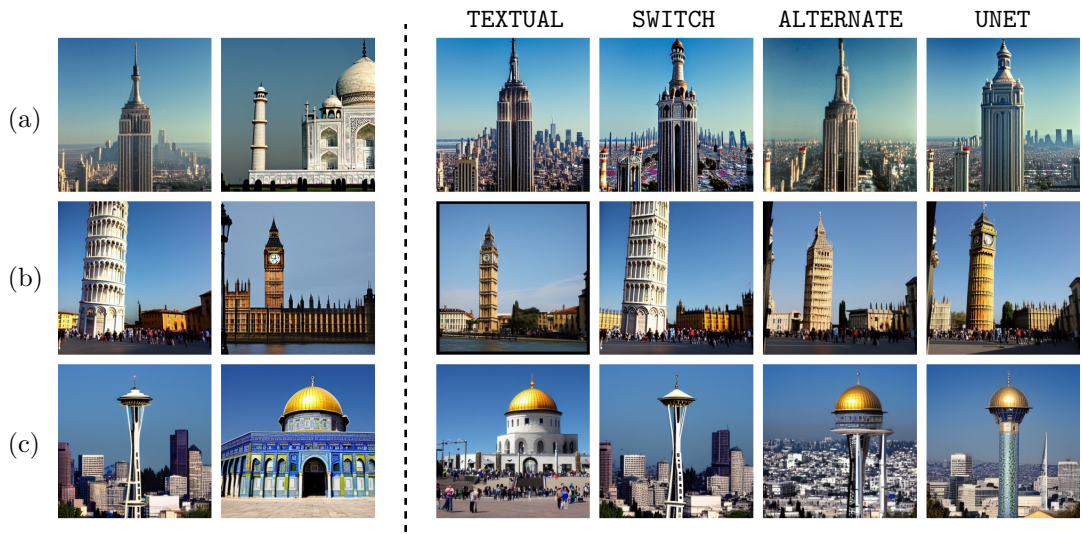

Building on our explorations in style (Section~\ref{subsec:paintings}) and emotion (Section~\ref{subsec:emotions}), we now turn to blending two distinct architectural concepts. This task is arguably more challenging than the previous ones, given the difficulty of faithfully reproducing complex geometries and fine architectural details \cite{style-cross-image}. For this reason, we selected iconic landmarks that feature distinct shapes and enjoy international recognition. From a conceptual blending perspective, each landmark contributes its defining structural features, which the model attempts to combine into a single, novel architectural entity.

We selected iconic landmarks such as the ``\emph{Empire State Building}'', ``\emph{Taj Mahal}'', ``\emph{Leaning Tower of Pisa}'', ``\emph{Big Ben}'', ``\emph{Seattle Space Needle}'', and ``\emph{Dome of the Rock}'', all notable for their recognizable shapes and distinct architectural cues, providing a test of the model's capacity to integrate complex geometry. 
Figure~\ref{fig:architecture} presents examples of these architectural blends. Here, we can see how \ALT and \UNE tend to yield more compelling fusions of structural forms, while \PRO typically fails to capture the intended overlap, and \SWI often produces simplified or underspecified outputs.
An illustrative case is blending the ``\emph{Leaning Tower of Pisa}'' with ``\emph{Big Ben}''. The \UNE method creates a rounded, tilted tower reminiscent of Pisa's hallmark shape, combined with a clock face and colour palette influenced by Big Ben, thereby preserving key features from both landmarks. In contrast, \PRO's latent-space interpolation rarely retains the spatial cues necessary to convey two overlapping architectures, and \SWI can suffer from a lack of detail when switching prompts mid-generation. Overall, these findings show that zero-shot architectural blending remains a difficult challenge for a simple stable diffusion backbone, yet methods like \UNE and \ALT can still produce convincing hybrids of iconic structures.

%% file: sections/validation-and-results.tex


\begin{wraptable}[36]{r}{0.48\textwidth}
\centering
\small
\caption{
    Full set of concept pairs organized into five categories, each illustrating a distinct type of blending scenario.
    The author of the painting in the Style section is omitted to better fit the table layout.
}
\label{tab:prompts_jair}
\begin{tabular}{ll}
    \toprule

    \rowcolor{gray!25}
    \multicolumn{2}{l}{\textbf{Same}} \\
    \midrule
    \emph{lion}      & \emph{cat} \\
    \emph{owl}       & \emph{tiger} \\
    \emph{avocado}   & \emph{pineapple} \\
    \emph{rabbit}    & \emph{lion} \\
    \emph{apple}     & \emph{hamburger} \\
    
    \midrule
    \rowcolor{gray!25}
    \multicolumn{2}{l}{\textbf{Different}} \\
    \midrule
    \emph{turtle}    & \emph{broccoli} \\
    \emph{blimp}     & \emph{whale} \\
    \emph{banana}    & \emph{shoes} \\
    \emph{canoe}     & \emph{walnuts} \\
    \emph{rocket}    & \emph{carrot} \\
    
    \midrule
    \rowcolor{gray!25}
    \multicolumn{2}{l}{\textbf{Compound Words}} \\
    \midrule
    \emph{butter}    & \emph{fly} \\
    \emph{bull}      & \emph{pit} \\
    \emph{kung fu}   & \emph{panda} \\
    \emph{tea}       & \emph{pot} \\
    \emph{man}       & \emph{bat} \\
    
    \midrule
    \rowcolor{gray!25}
    \multicolumn{2}{l}{\textbf{Concept and Style}} \\
    \midrule
    \emph{A portrait of a man} & \emph{Girl with a Pearl Earring} \\
    \emph{A portrait of a man} & \emph{Guernica} \\
    \emph{A portrait of a man} & \emph{The Scream} \\
    \emph{A portrait of a man} & \emph{Starry Night} \\
    
    \midrule
    \rowcolor{gray!25}
    \multicolumn{2}{l}{\textbf{Architecture}} \\
    \midrule
    \emph{Duomo of Milan} & \emph{Taj Mahal} \\
    \emph{Eiffel Tower} & \emph{Sagrada Familia} \\
    \emph{Leaning Tower of Pisa}   & \emph{Big Ben} \\
    
    \bottomrule
\end{tabular}
\end{wraptable}

The quality of any conceptual blend often depends on the nature of the concepts being combined. To comprehensively evaluate our blending strategies, we tested each method on five distinct categories of concept pairs, as summarized in Table~\ref{tab:prompts_jair}. These categories are intended to highlight divergent properties of the concepts and the resulting blends.
Each category represents a different blending scenario. When two concepts belong to the same class, such as pairs of animals (e.g., ``\emph{lion-cat}'') or foods (``\emph{avocado-pineapple}''), they typically share a visual or semantic common ground that can facilitate a more coherent blend. Conversely, different class objects (e.g., ``\emph{turtle-broccoli}'') often introduce stronger contrasts in shape, texture, or context, potentially challenging the blending process. Compound words, such as ``\emph{butter-fly}'' or ``\emph{tea-pot}'', offer further complexity: many of these compounds merge two distinct concepts into a single expression (e.g., ``\emph{butterfly}''), raising the question of whether the blending method will capture the conventional meaning of the compound or produce a literal hybrid of its components. We also examine concept and style, pairing a descriptive prompt like ``\emph{a portrait of a man}'' with a specific artistic style (e.g., ``\emph{Picasso}'', ``\emph{van Gogh}''), a task reminiscent of style transfer but without any specialized fine-tuning. Finally, we investigate architecture, blending visually iconic structures (e.g., ``\emph{Duomo-Taj Mahal}'', ``\emph{Pisa-Big Ben}'') to see how these methods handle complex geometric and structural features.

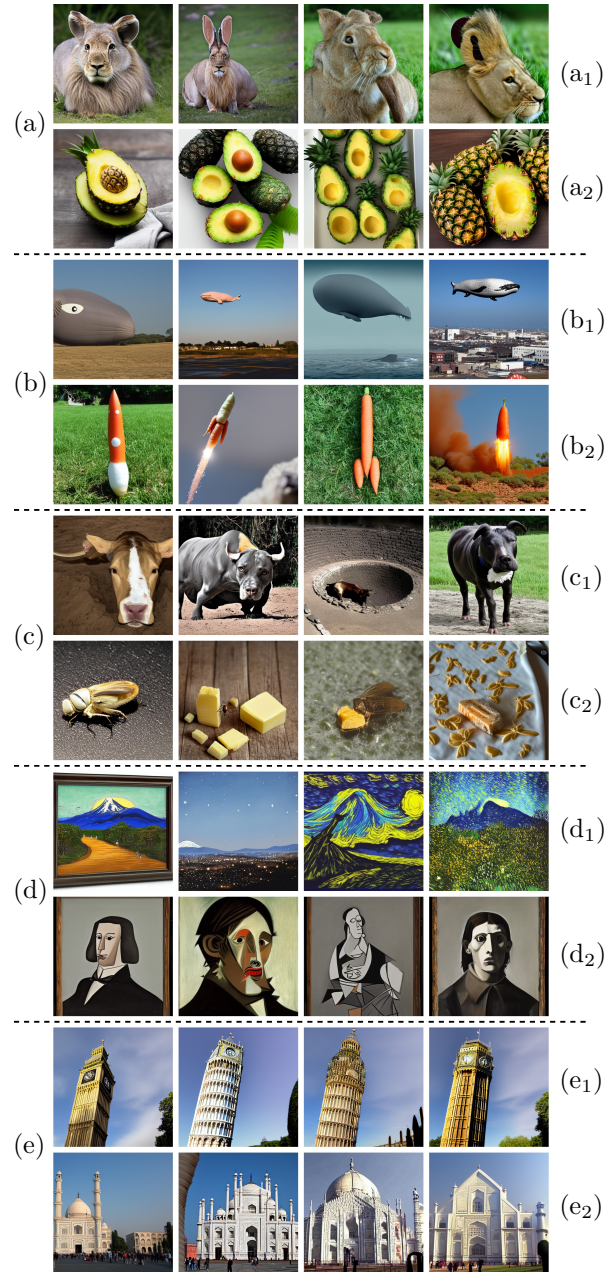
\begin{wrapfigure}[42]{R}{0.45\textwidth}
    \centering
    \begin{overpic}[
      percent,
      grid=false,
      tics=10,
      scale=0.5,
      width=0.5\textwidth
      ]{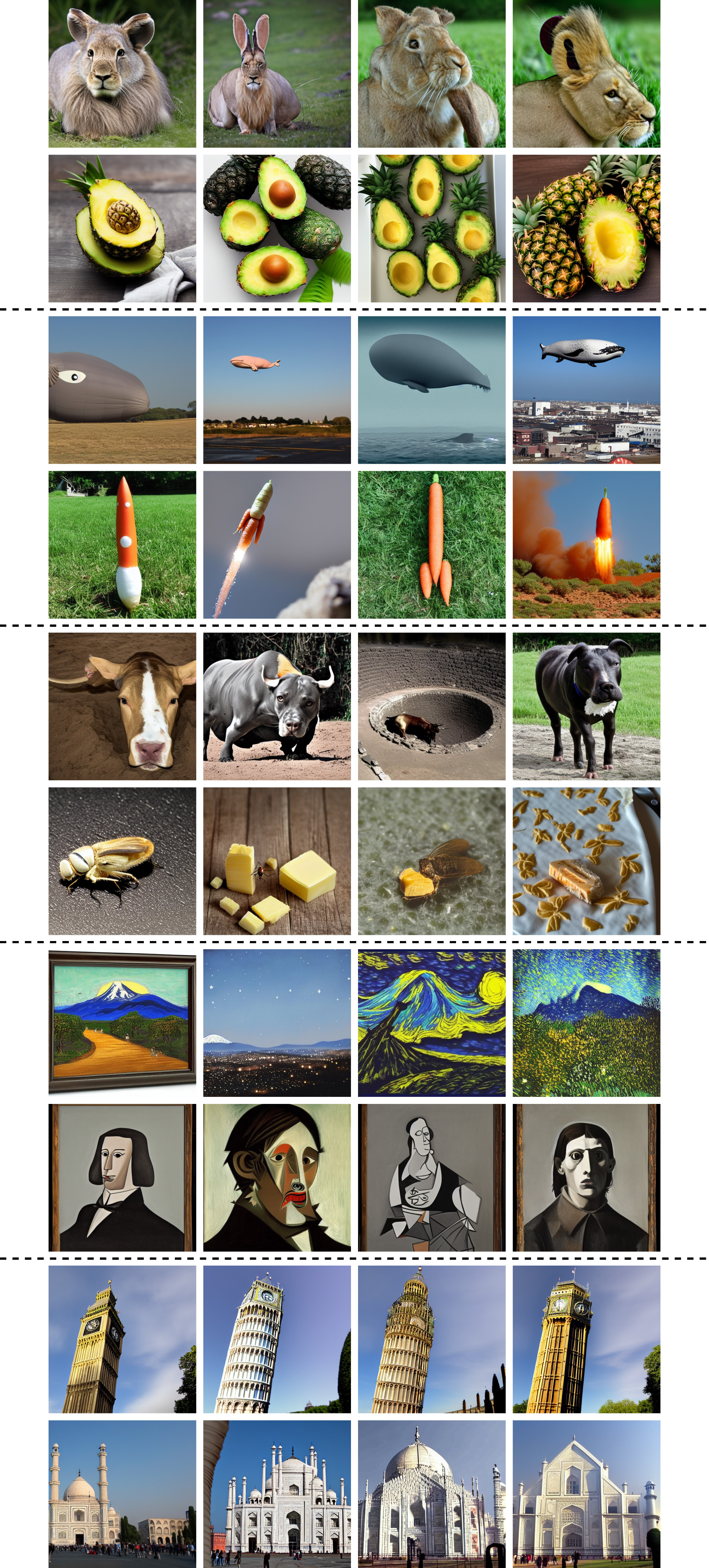}

      \put(-0.1, 90){\small(a)}
      \put(43, 94){\small($\text{a}_1$)}
      \put(43, 85){\small($\text{a}_2$)}

      \put(-0.1, 69.5){\small(b)}
      \put(43, 74.5){\small($\text{b}_1$)}
      \put(43, 64.5){\small($\text{b}_2$)}

      \put(-0.1, 49.5){\small(c)}
      \put(43, 54.2){\small($\text{c}_1$)}
      \put(43, 44.5){\small($\text{c}_2$)}

      \put(-0.1, 29.5){\small(d)}
      \put(43, 34.3){\small($\text{d}_1$)}
      \put(43, 24.4){\small($\text{d}_2$)}

      \put(-0.1, 9.3){\small(e)}
      \put(43, 14.5){\small($\text{e}_1$)}
      \put(43, 4.5){\small($\text{e}_2$)}

    \end{overpic}
   \caption{
        Two representative blends from each category (a)--(e), showcasing variations within pairs ($\text{a}_1$, $\text{a}_2$) to ($\text{e}_1$, $\text{e}_2$). Highlights key characteristics of the user study materials.
    }
    \label{fig:survey-sample-images}
\end{wrapfigure}

Because the notion of a ``successful blend'' is inherently subjective, the evaluation was conducted through a user study. Participants were shown images generated by different blending methods for the same concept pair, and they were asked to rank each image according to how well it captured a coherent hybrid of the two prompts. Figure~\ref{fig:survey-sample-images} illustrates a subset of these images and identifies which blending method generated them. The same set of five categories and prompts (Table~\ref{tab:prompts_jair}) was presented, with the main focus on comparing how each method performed under these different blending scenarios.

\newpage
\subsection{User Study}


A user study was conducted with 100 participants, all young adults between 19 and 25 years of age, to evaluate the performance of the four blending methods introduced earlier. Each participant was presented with 22 pairs of prompts, yielding a total of 84 images distributed across the five previously described categories. Before the study, participants received a brief 10-minute introductory session explaining the notion of concept blending. They were then asked to rank each displayed image according to how effectively it conveyed a coherent blend of the two prompts.
All participants were informed of the prompt pairs under consideration (e.g., ``\emph{lion-cat}'', ``\emph{turtle-broccoli}'', ``\emph{a portrait of a man-Guernica by Pablo Picasso}''), but the particular blending method that generated each image was withheld to avoid bias. Moreover, images were presented in a randomized order for each participant and for each concept pair, further minimizing any systematic preference or learning effects over time.

Since there is no single ``ground truth'' definition of what constitutes a ``successful'' conceptual blend, subjective human judgment is essential \cite{MUPLC15}. Conceptual blending inherently involves creativity and can therefore be interpreted differently by different people. By collecting participant rankings, we obtain a direct measure of which blends best capture an intended combination of ideas.

As discussed in Section~\ref{sec:seed-dependency}, the outputs of diffusion models, and by extension also our methods, are highly dependent on the seed used during sampling. To minimize the impact of this variability, we manually selected the most appealing result for each method from a pool of 10 predefined seeds. Additionally, each blend was created using an equal ratio of the two prompts (see Section~\ref{subsec:blend_ratio}), preventing the synthesis process from favouring one concept over the other. Although this procedure could, in some cases, yield suboptimal blends, it keeps the experimental conditions balanced across all methods and prompt pairs.

\subsection{Methodology for Evaluation}

In total, 100 participants were asked to rank the results of all four blending
methods over the 22 pairs of concepts. Our goal is to understand which,
if any, of the methods is preferred over this sample.
To achieve this, we analyse the relative preference of each method over all
others throughout the full survey. In other words, we analyse the frequency at 
which one method is preferred over another. The rationale behind this
choice is the following: if two methods are equally
effective, since participants are required to provide relative 
rankings, it is reasonable to expect that half of the participants ``prefer'' one of the methods, and half of them ``prefer'' the other.
The preference of a method over another can be modelled as a Bernoulli
distribution, where the parameter represents the probability that
a method is ranked better than another one. Our goal is to verify whether this parameter is close to 0.5 (i.e., there is no evidence that the subjects show a preference for one method) or away from this 
middle point, indicating a preference in one direction or the other.

For each possible pair of methods \texttt{METHOD-1}, \texttt{METHOD-2}, we estimate the Bernoulli parameter by counting
the proportion of answers in which \texttt{METHOD-1} has a lower rank than (i.e., it is \emph{preferred over}) 
\texttt{METHOD-2}. Table~\ref{tab:survey-results} shows the relations where this proportion is above 0.5,
suggesting a direct preference among the participants of the survey. 
This induces a partial order. For instance, the first row of Table~\ref{tab:survey-results}
expresses that \ALT was preferred over \SWI, in 51\% of all the comparisons. While this expresses \emph{some} evidence in favour of
\ALT, it is not clear whether this has any statistical power. Hence,
the last column of the table presents the p-value for the null hypothesis $H_0: \ALT>\SWI$. We use these p-values to verify whether 
the preference is statistically significant. In the table, ${\sim} 0$ expresses
that the p-value is below $0.001$, which can be interpreted as being \emph{extremely} significant.%
\footnote{We consider evidence to be \emph{statistically significant} if the p-value is $<0.05$; \emph{very significant} if it is $<0.01$; and \emph{extremely significant} if below $0.001$.}
\begin{table}[t!]
  \centering
  \caption{
    Pair-wise comparison of user preferences of methods, and their p-value.
    ${\sim} 0$ expresses that the p-value is below $0.001$.
  }
  \begin{tabular}{@{}lcc@{}}
    \toprule
    \textbf{Preference}	& \textbf{Proportion}	& \textbf{p-value} \\ \midrule
    $\ALT < \SWI$		& 0.51			& 0.31 \\ 
    $\ALT < \PRO$		& 0.60			& ${\sim}0$ \\ 
    $\ALT < \UNE$		& 0.57 			& ${\sim}0$ \\ 
    $\SWI < \PRO$		& 0.61 			& ${\sim}0$ \\ 
    $\SWI < \UNE$		& 0.58			& ${\sim}0$ \\ 
    $\UNE < \PRO$		& 0.55			& ${\sim}0$ \\ 
    \bottomrule
  \end{tabular}
  \label{tab:survey-results}
\end{table}
%

\begin{figure}
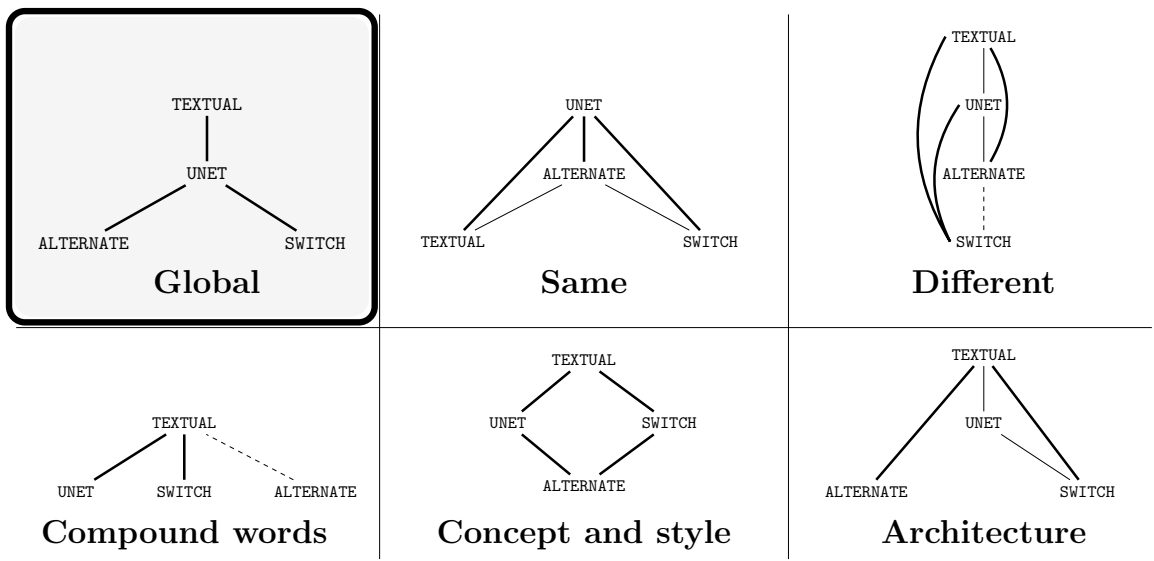

\includestandalone[width=\textwidth]{images/fullhasse2}
\caption{Ordering preferences by category. In the Hasse diagrams, thick lines mean extremely significant 
(p-value ${} < 0.001$), thin lines mean very significant (p-value ${} < 0.01$) and dashed lines mean significant
(p-value ${} < 0.05$).}
\label{fig:comp:categories}
\end{figure}

In summary,
the first row in Table~\ref{tab:survey-results} indicates that, although
more participants ranked \ALT better than \SWI, there is no statistical evidence
to conclude that \ALT is preferred over \SWI. However, the subsequent rows
respectively demonstrate a clear preference for both \ALT and \SWI over \PRO and \UNE.
Furthermore, the data shows a preference for \UNE over \PRO, with all these
outcomes exhibiting a high degree of statistical significance (p-value
essentially 0). In other words, the data indicates a tie between \ALT and \SWI,
followed by a preference for both over \UNE, and of the latter over \PRO. These preference relations are
summarised in the grey box with bold border of Figure~\ref{fig:comp:categories}: methods depicted lower in the structure are more preferred than those 
higher above; the lack of connection between \ALT and \SWI represents the missing evidence in favour of one over the other.
%
%

We repeat
the same analysis divided by type, as shown in
Table~\ref{tab:comparison-results-categories} and the remaining blocks
of Figure~\ref{fig:comp:categories}.
\begin{table}[t!]
  \caption{
    Pair-wise comparison of user preferences of methods, and their p-value, grouped by the type of blend executed.
    ${\sim} 0$ expresses that the p-value is below $0.001$.
  }
  \label{tab:comparison-results-categories}
  \centering
  \begin{tabular}{@{}clcc@{}}
    \toprule
    \textbf{Cat.} & \textbf{Test} & \textbf{Proportion} & \textbf{p-value} \\
    \midrule
    \multirow{6}{*}{\rotatebox{90}{Same}}
                      & $\SWI < \ALT$	& 0.57	& 0.002 \\
                      & $\PRO < \ALT$	& 0.57 	& 0.001 \\
                      & $\ALT < \UNE$	& 0.60	& ${\sim}0$ \\
                      & $\SWI < \PRO$	& 0.51	& 0.33 \\
                      & $\SWI < \UNE$	& 0.69	& ${\sim}0$ \\
                      & $\PRO < \UNE$ & 0.70	& ${\sim}0$ \\
    \midrule
    \multirow{6}{*}{\rotatebox{90}{Different}}
                      & $\SWI < \ALT$ & 0.54 & 0.047 \\
                      & $\ALT < \PRO$ & 0.62 & ${\sim}0$ \\
                      & $\ALT < \UNE$ & 0.57 & 0.001 \\
                      & $\SWI < \PRO$ & 0.59 & ${\sim}0$ \\
                      & $\SWI < \UNE$ & 0.59 & ${\sim}0$ \\
                      & $\UNE < \PRO$ & 0.54 & 0.038 \\
    \midrule
    \multirow{6}{*}{\rotatebox{90}{\parbox[c]{2cm}{\centering Compound \\ words}}}
                      & $\ALT < \SWI$ & 0.51 & 0.269 \\
                      & $\ALT < \PRO$ & 0.55 & 0.021 \\
                      & $\UNE < \ALT$ & 0.54 & 0.056 \\
                      & $\SWI < \PRO$ & 0.62 & ${\sim}0$ \\
                      & $\SWI < \UNE$ & 0.53 & 0.079 \\
                      & $\UNE < \PRO$ & 0.62 & ${\sim}0$ \\
    \midrule
    \multirow{6}{*}{\rotatebox{90}{\parbox[c]{2cm}{\centering Concept  {} \\ and Style}}}
                      & $\ALT < \SWI$ & 0.67 & ${\sim}0$ \\
                      & $\ALT < \PRO$ & 0.82 & ${\sim}0$ \\
                      & $\ALT < \UNE$ & 0.69 & ${\sim}0$ \\
                      & $\SWI < \PRO$ & 0.66 & ${\sim}0$ \\
                      & $\UNE < \SWI$ & 0.50 & 0.44 \\
                      & $\UNE < \PRO$ & 0.76 & ${\sim}0$ \\
    \midrule
    \multirow{6}{*}{\rotatebox{90}{Architecture}}
                      & $\SWI < \ALT$	& 0.54 & 0.086 \\
                      & $\ALT < \PRO$ & 0.62 & ${\sim}0$ \\
                      & $\ALT < \UNE$ & 0.54 & 0.086 \\
                      & $\SWI < \PRO$ & 0.69 & ${\sim}0$ \\
                      & $\SWI < \UNE$ & 0.57 & 0.008 \\
                      & $\UNE < \PRO$ & 0.59 & 0.002 \\
    \bottomrule
  \end{tabular}
\end{table}
A simple analysis of the figures---where dashed lines express baseline significance, thin lines high significance and thick lines extreme significance---shows that in general \ALT and \SWI are the most preferred methods, while \PRO is the least preferred with \UNE remaining in the middle. This trend is, however, reversed for
\emph{Same}, where no method is preferred over \PRO. We also see that
preferences are indeed dependent on the specific category of blending tested, with no two categories producing the same preference relation.


%
\begin{table}[tb]
\caption{Statistics about the rank of each method over the whole survey, divided by pairs of concepts and category.}
\label{tab:means}
\resizebox{\textwidth}{!}{%

\includestandalone{fulltable}

}
\end{table}
%
Moreover, in Table~\ref{tab:means}, we show the mean (rounded to three significant digits), median, and mode rank assigned to each of the four methods, divided by individual pairs of concepts and by category. 
The last line 
also summarises the results globally. The mode tells us the rank that was selected the most, while the median expresses to which rank were at least half of the answers assigned. Hence for instance in the first row (\emph{lion-cat}), \SWI is ranked most often in first place (mode 1) and half of the participants rank it within the first 2 places (median 2). These numbers confirm the previous statistical analysis.
Indeed, \ALT and \SWI tend to be ranked at a better position than \UNE, and \PRO is typically evaluated last. The exception is in the category \emph{Same}, where \PRO presents mode 1 and median 2 in its rank. We can see that this behaviour is highly influenced by the pair
\emph{rabbit-lion} in which more than half of the participants rank \PRO as the best method.
It is also evident from the table that neither \ALT nor \SWI clearly outperforms the other: although \ALT generally has a lower mode (its most frequently assigned rank), \SWI often shows a lower median, indicating that more participants place it in a higher position overall.








\subsection{Discussion}\label{sec:discussion}
The comparison of the methods in the user survey, provides a clear hierarchy of
preferences supported by a formal statistical analysis: \ALT{} and \SWI{} are equally liked, and preferred over \UNE, which is itself preferred over \PRO{}. In this section, we discuss the strengths and weaknesses of
these methods explaining the results of the user survey.

\begin{figure}[t!]
    \centering
    \begin{overpic}[percent, grid=false, tics=10, scale=.5, width=0.89\linewidth]
        {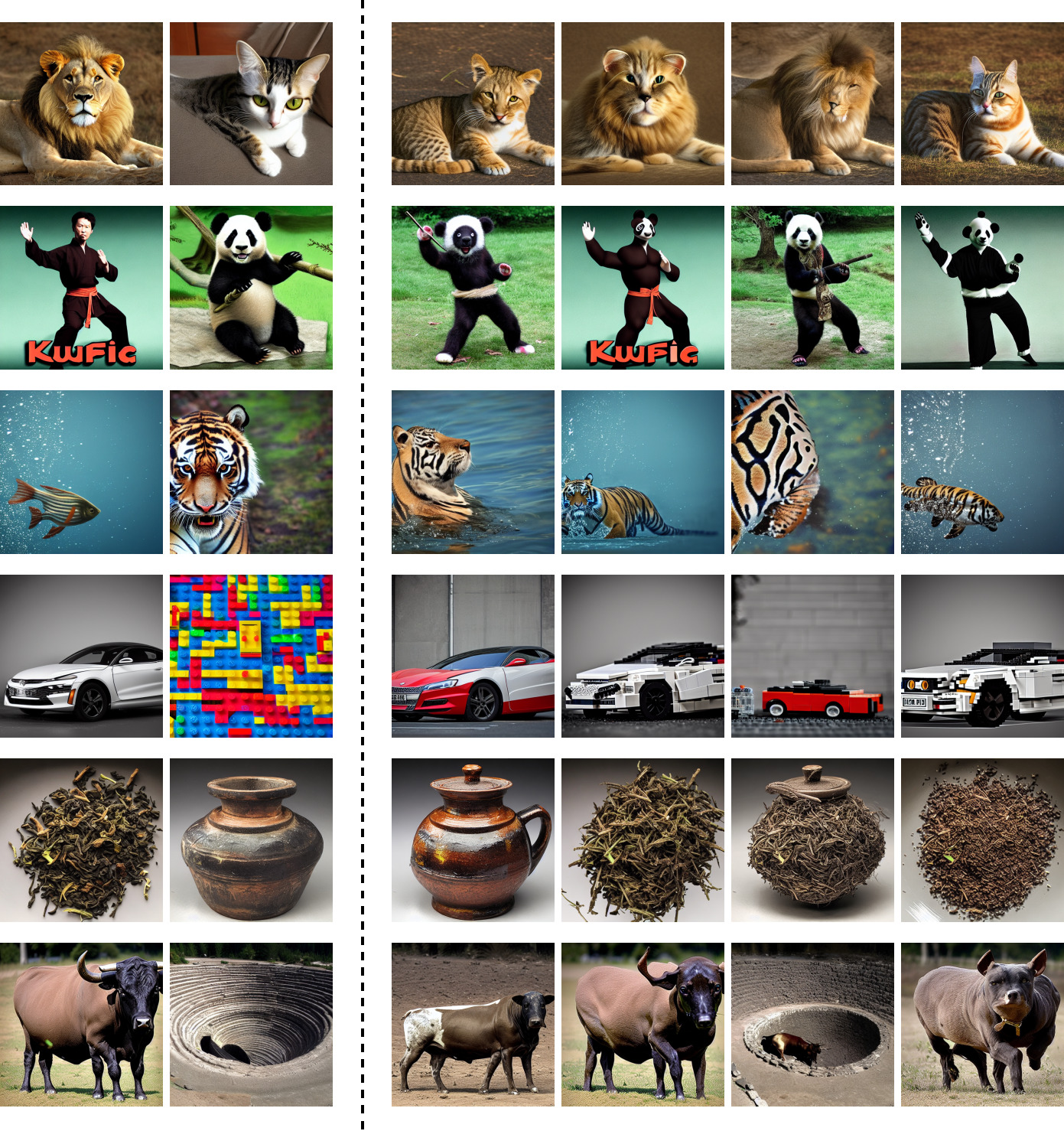}
        \put(37.5, 98.8){\small\PRO}
        \put(53.1, 98.8){\small\SWI}
        \put(66.1, 98.8){\small\ALT}
        \put(84, 98.8){\small\UNE}

        \put(-4.0, 90){\small(a)}
        \put(-4.0, 74){\small(b)}
        \put(-4.0, 58){\small(c)}
        \put(-4.0, 41){\small(d)}
        \put(-4.0, 25.5){\small(e)}
        \put(-4.0, 9){\small(f)}
    \end{overpic}
    \caption{Comparison of the blending methods with the same seed. On the left,
    the individual prompts, on the right, the blended images.
        (a)~``\emph{lion-cat}'', (b)~``\emph{kung fu-panda}'', (c)~``\emph{fish-tiger}'',
        (d)~``\emph{car-lego bricks}'', (e)~``\emph{tea-pot}'', and
        (f)~``\emph{bull-pit}''.
    }\label{fig:same-seed}
\end{figure}
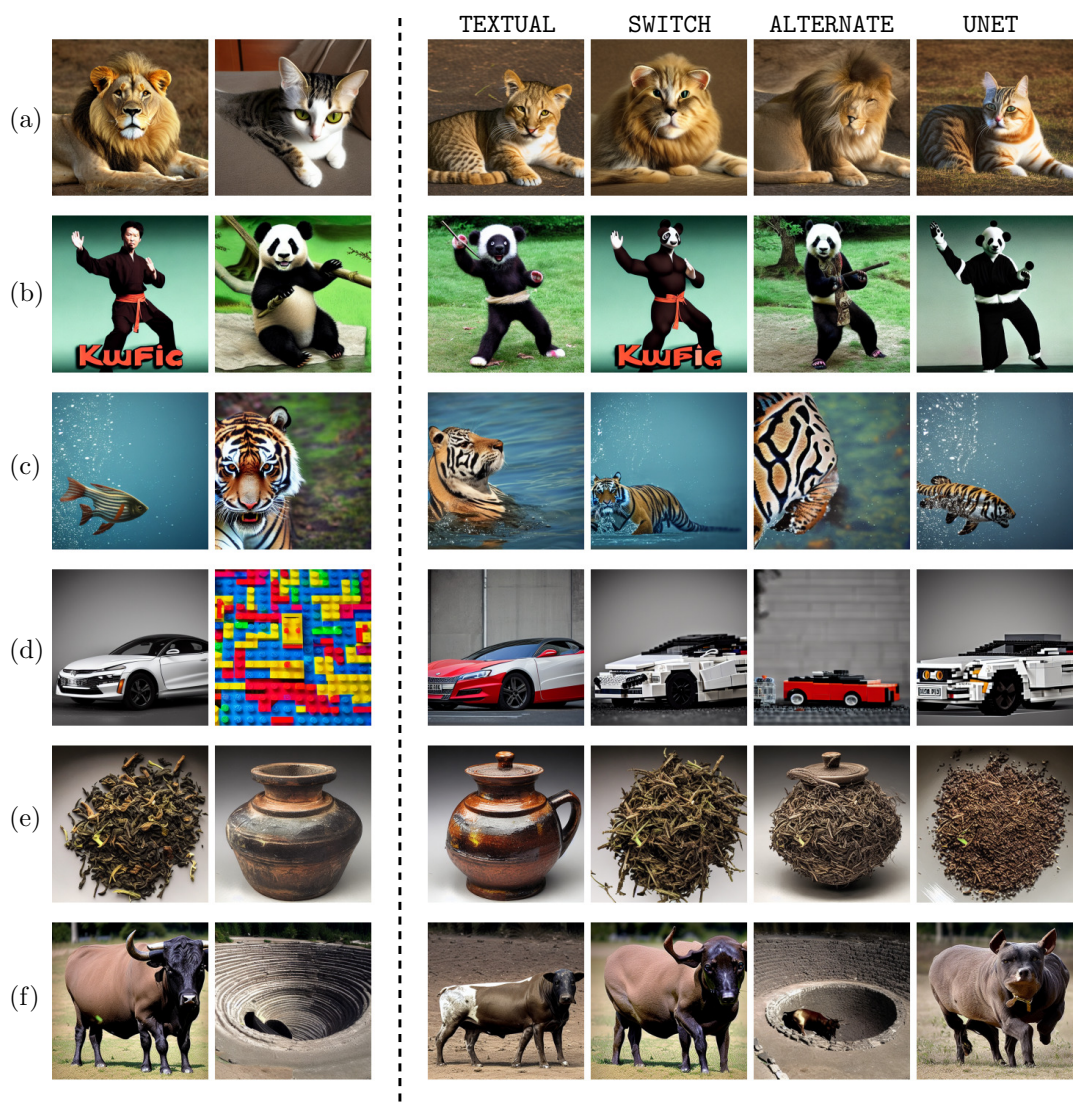

Figure~\ref{fig:same-seed} shows a series of samples generated by blending the
following prompt pairs: ``\emph{lion-cat}'', ``\emph{kung fu-panda}'',
``\emph{fish-tiger}'', ``\emph{car-lego bricks}'', ``\emph{tea-pot}'', and
``\emph{bull-pit}''. To focus on the specific characteristics of each method,
these images are generated starting from the same initial noise. Moreover, analogous to the images used in the user study, the blending ratio was kept constant at $0.5$ across all methods.

\paragraph{Blending in the Textual Latent Space (\PRO).}
\label{par:blending-latent-space}
\PRO’s interpolation in the latent space of prompts does not always produce meaningful visual hybrids, as evidenced by the method’s consistently lower ranking in the user survey. We attribute this shortcoming to how the text encoder (CLIP in Stable Diffusion v1.4) arranges prompts in its latent space \cite{clip,stable-diffusion}. There is no guarantee that points interpolated between two concept embeddings correspond to semantically valid blends; for instance, \emph{``car''} and \emph{``lego bricks''} in Figure~\ref{fig:same-seed} yield a mostly unaltered vehicle with few, if any, Lego-like details. Moreover, we observe substantial seed-to-seed variability, so even if the text prompt embedding remains constant, small differences in the initial noise can affect how visual and semantic elements are merged. Nonetheless, \PRO sometimes outperforms other methods for concept pairs in the \emph{Same} category (e.g., closely related items), likely because the overlap in visual or semantic features is easier to capture via simple interpolation. 

\paragraph{Blending in the Iterative Reverse Process (\SWI and \ALT).}
The \SWI{} and \ALT{} methods blend two concepts by altering the
conditioning prompt during the iterative denoising process of the diffusion
pipeline. These two methods are tied as the most effective overall, with \ALT{},
despite its simplicity, showing a slight advantage in \emph{Different}
and \emph{Architecture} contexts. This indicates that while both methods are
generally reliable, the choice between them may be influenced by the specific
blending context.

As mentioned in Section~\ref{sec:switch-pipeline}, results from \SWI{} vary
considerably depending on the timestep at which the prompt is switched from one concept to the other. Finding
the right timestep is crucial for achieving a good visual blend. This behaviour is
particularly evident in the case of ``\emph{tea-pot}'', as illustrated in
Figure~\ref{fig:same-seed}. While ``\emph{tea}'' and ``\emph{pot}'' together form a
semantically meaningful compound word, they represent visually distinct
concepts. Recall that \SWI starts by generating an image based on the
first prompt (in this case ``\emph{tea}''), and then---midway through the diffusion process---switches to the second prompt (``\emph{pot}''). 
After the switch, the model
struggles to change and adjust the existing distribution to align with the prompt \emph{pot}. Hence, the final image predominantly depicts the first
concept, failing to blend the two effectively.

Another undesired behaviour of \SWI{} is the \emph{cartoonification} of the
produced blend. When unable to shift the pixel
distribution towards the new prompt, the diffusion pipeline corrects the existing noisy image latent by
progressively removing the high-frequency details, resulting in a picture with
flat colours and simplified shapes. This behaviour can be observed in
the ``\emph{kung fu-panda}'' blend produced by \SWI{} in Figure~\ref{fig:same-seed}.
In all our experiments, this kind of behaviour could not be observed in any of the other methods.

The \ALT{} method, which alternates between the two prompts at each timestep,
tends to produce consistent results when the two blended concepts are visually
distant. The unique type of visual
blend that this technique produces is arguably even more interesting when the two concepts are both visually and
semantically very different. This is the case of ``\emph{tea-pot}'' and
``\emph{butter-fly}'', where the model creates an image that literally and
spatially contains both the first and the second prompt. This is also evident in
the ``\emph{bull-pit}'' blend in Figure~\ref{fig:survey-sample-images}, where the
\ALT{} generates what could be described as a \emph{a bull in a pit}.

\paragraph{Blending in the U-Net (\UNE).}
\UNE{} occupies a middle ground in the rankings, consistently
outperforming \PRO{} but falling behind both \ALT{} and \SWI{} in most
comparisons. Its performance indicates that it is a viable blending method in
certain contexts but lacks the robustness and adaptability of the top-performing
methods.
Compared to the other blending methods, \UNE{}, which encodes the image (latent)
conditioned on the first prompt in the bottleneck and then decodes it with the
second prompt, produces more subtle blends while maintaining generally
consistent results. This subtlety might explain why \UNE{} ranks below \ALT{}
and \SWI{} in our user study, as the visual blend is less pronounced compared to
the more evident combinations achieved by the other methods.

From our experiments, in general, \UNE{} seems to be able to modify the shape of
the first concept with prominent features of the second. This is particularly evident in
the ``\emph{fish-tiger}'' and ``\emph{car-lego bricks}'' blends shown in
Figure~\ref{fig:same-seed}. Interestingly, on the ``\emph{kung fu-panda}''
blend, \UNE{} seems to slightly change the visual representation of the first
prompt while matching the colours of the second one. This subtle blending is also
evident in the ``\emph{bull-pit}'' blend of Figure~\ref{fig:same-seed}, where
surprisingly the pipeline creates an image that somewhat resembles
an actual \emph{pitbull} although nothing in the prompt makes any 
reference to this breed of dog.

Although the \UNE{} method consistently produces more subtle blends, it relies
on the alignment of the visual depiction of the first prompt with the embedding
of the second. Specifically, the image generated by the diffusion pipeline based
on the first prompt must possess visual features that can be effectively
leveraged by the cross-attention layers in the U\mbox{-}Net when incorporating the
embedding of the second prompt. When this is not the case, like in
the case of blending ``\emph{tea}'' and ``\emph{pot}'', the model cannot
identify which elements of the first prompt should be modified by the second one, thus producing unappealing results.

Although the \UNE{} method consistently produces more subtle blends, its results
depend on the alignment between the visual representation of the first prompt
and the embedding of the second. Specifically, the image generated by the
diffusion pipeline based on the first prompt must contain visual features that
can be effectively leveraged by the cross-attention layers in U-Net when
incorporating the embedding of the second prompt. When this alignment is not
possible, as in the case of the blend between ``\emph{tea}'' and ``\emph{pot}'', the
model struggles to identify which elements of the first prompt should be
modified by the second, resulting in an unsuccessful blend.

%% file: images/fullhasse2.tex
\begin{tikzpicture} 
\begin{scope}[xshift=-8.3cm]
\node (u) {\scalebox{1.1}{\UNE}};
\node[below left=of u] (a) {\scalebox{1.1}{\ALT}};
\node[below right=of u] (s) {\scalebox{1.1}{\SWI}};
\node[above=of u] (p) {\scalebox{1.1}{\PRO}};
\node[above=of p] (null) {};
\draw[ultra thick] (a) -- (u) -- (p) (s) -- (u);
\node[anchor=north] at (0,-2) (n) {\scalebox{1}{\bf\huge Global}};
\begin{scope}[on background layer]
\node[rectangle,fill=black!4,fit=(a)(s)(p)(u)(n)(null),rounded corners=10,draw=black!4,inner sep=10,ultra thick] (box) {};
\node[draw=black,fit=(box),line width=4,rounded corners=10] {};
\end{scope}
\end{scope}
\node (u) {\ALT};
\node[below left=of u] (a) {\PRO};
\node[below right=of u] (s) {\SWI};
\node[above=of u] (p) {\UNE};
\draw (a) -- (u) (s) -- (u);
\draw[ultra thick] (u) -- (p) (a) -- (p) (s) -- (p);
\node[anchor=north] at (0,-2) (n) {\bf\huge Same};

\begin{scope}[xshift=8.8cm]
\node (u) {\ALT};
\node[below=of u] (a) {\SWI};
\node[above=of u] (s) {\UNE};
\node[above=of s] (p) {\PRO};
\draw[dashed] (a) -- (u);
\draw (u) -- (s) (s) -- (p);
\draw[ultra thick] (a.west) to[bend left] (s.west) (a.west) to[bend left] (p.west) (u) to[bend right] (p);
\node[anchor=north] at (0,-2) {\bf\huge Different};
\end{scope}

\begin{scope}[xshift=-8.8cm,yshift=-5.5cm]
\node (u) {\PRO};
\node[below=of u] (a) {\SWI};
\node[below left=of u] (s) {\UNE};
\node[below right=of u] (p) {\ALT};
\draw[dashed] (p) -- (u);
\draw[ultra thick] (s) -- (u) (a) -- (u);
\node[anchor=north] at (0,-2) {\bf\huge Compound words};
\end{scope}

\begin{scope}[xshift=0cm,yshift=-5.5cm]
\node (c) {};
\node[left=of c] (u) {\UNE};
\node[below=of c] (a) {\ALT};
\node[right=of c] (s) {\SWI};
\node[above=of c] (p) {\PRO};
\draw[ultra thick] (a) -- (u) -- (p) -- (s) -- (a);
\node[anchor=north] at (0,-2) {\bf\huge Concept and style};
\end{scope}

\begin{scope}[xshift=8.8cm,yshift=-5.5cm]
\node (u) {\UNE};
\node[below left=of u] (a) {\ALT};
\node[below right=of u] (s) {\SWI};
\node[above=of u] (p) {\PRO};
\draw (s) -- (u) -- (p);
\draw[ultra thick] (a) -- (p) (s) -- (p);
\node[anchor=north] at (0,-2) {\bf\huge Architecture};
\end{scope}

\begin{scope}[on background layer]
\draw (-12.5,-3.4) -- (12.5,-3.4);
\draw (-4.5,3.5) -- (-4.5,-8.5);
\draw (4.5,3.5) -- (4.5,-8.5);
\end{scope}

\end{tikzpicture}

%% file: fulltable.tex
\begin{tabular}{@{}cc*{4}{>{\columncolor{lightgray!60}}c>{\columncolor{lightgray!30}}cc}@{}}
	\toprule
	                                 & \multirow{2}{*}{Prompt} & \multicolumn{3}{c}{\ALT} & \multicolumn{3}{c}{\SWI} & \multicolumn{3}{c}{\PRO} & \multicolumn{3}{c}{\UNE}                                                               \\
	                                 &                         & mean                     & median                   & mode                     & mean                     & median & mode & mean & median & mode & mean & median & mode \\
	\midrule
	\multirow{6}{*}{\rotatebox{90}{\scalebox{1}{Same}}}
	                                 & Lion-Cat                & 2.70                     & 3                        & 4                        & 2.02                     & 2      & 1    & 2.61 & 3      & 4    & 2.67 & 3      & 3    \\
	                                 & Owl-Tiger               & 2.27                     & 2                        & 1                        & 1.87                     & 2      & 1    & 2.30 & 2      & 3    & 3.55 & 4      & 4    \\
	                                 & Avocado-Pineapple       & 2.02                     & 2                        & 2                        & 2.53                     & 3      & 2    & 2.43 & 3      & 1    & 3.02 & 3      & 4    \\
	                                 & Rabbit-Lion             & 3.27                     & 4                        & 4                        & 2.41                     & 2      & 2    & 1.59 & 1      & 1    & 2.73 & 3      & 3    \\
	                                 & Apple-Hamburger         & 2.43                     & 2                        & 2                        & 2.32                     & 2      & 2    & 2.26 & 2      & 3    & 2.99 & 3      & 4    \\
	\cmidrule{2-14}
	                                 & CATEGORY TOTAL          & 2.54                     & 3                        & 2                        & 2.23                     & 2      & 1    & 2.24 & 2      & 1    & 2.99 & 3      & 4    \\
	\midrule
	\multirow{6}{*}{\rotatebox{90}{Different}}
	                                 & Turtle-Broccoli         & 2.51                     & 3                        & 3                        & 2.48                     & 2      & 2    & 1.44 & 1      & 1    & 3.57 & 4      & 4    \\
	                                 & Blimp-Whale             & 2.75                     & 3                        & 3                        & 1.89                     & 2      & 1    & 3.37 & 4      & 4    & 1.99 & 2      & 2    \\
	                                 & Banana-Shoes            & 1.54                     & 1                        & 1                        & 1.86                     & 2      & 1    & 2.96 & 3      & 3    & 3.63 & 4      & 4    \\
	                                 & Canoe-Walnuts           & 1.72                     & 1                        & 1                        & 3.37                     & 4      & 4    & 2.58 & 3      & 2    & 2.33 & 2      & 3    \\
	                                 & Rocket-Carrot           & 3.20                     & 3                        & 3                        & 1.77                     & 2      & 2    & 3.44 & 4      & 4    & 1.59 & 1      & 1    \\
	\cmidrule{2-14}
	                                 & CATEGORY TOTAL          & 2.35                     & 2                        & 3                        & 2.27                     & 2      & 2    & 2.76 & 3      & 4    & 2.62 & 3      & 4    \\
	\midrule
	\multirow{6}{*}{\rotatebox{90}{\parbox[c]{2cm}{\centering Compound                                                                                                                                                                   \\ Words}}}
	                                 & Butter-Fly              & 2.48                     & 2                        & 2                        & 3.15                     & 4      & 4    & 2.24 & 2      & 1    & 2.14 & 2      & 1    \\
	                                 & Bull-Pit                & 3.14                     & 4                        & 4                        & 1.61                     & 1      & 1    & 2.92 & 3      & 3    & 2.33 & 2      & 2    \\
	                                 & Kung fu-Panda           & 1.80                     & 1                        & 1                        & 2.36                     & 2      & 2    & 3.40 & 4      & 4    & 2.45 & 3      & 3    \\
	                                 & Tea-Pot                 & 1.76                     & 2                        & 1                        & 2.85                     & 3      & 3    & 2.11 & 2      & 1    & 3.28 & 3      & 4    \\
	                                 & Man-Bat                 & 3.21                     & 3                        & 3                        & 1.83                     & 2      & 2    & 3.26 & 3      & 4    & 1.70 & 1      & 1    \\
	\cmidrule{2-14}
	                                 & CATEGORY TOTAL          & 2.48                     & 2                        & 1                        & 2.36                     & 2      & 2    & 2.79 & 3      & 3    & 2.38 & 2      & 1    \\
	\midrule
	\multirow{5}{*}{\rotatebox{90}{\parbox[c]{2cm}{\centering Concept {}                                                                                                                                                                 \\ and Style}}}
	                                 & Vermeer                 & 1.98                     & 2                        & 1                        & 2.27                     & 2      & 1    & 3.16 & 4      & 4    & 2.59 & 3      & 2    \\
	                                 & Picasso                 & 2.13                     & 2                        & 1                        & 2.05                     & 2      & 2    & 3.19 & 3      & 4    & 2.63 & 3      & 3    \\
	                                 & Munch                   & 1.67                     & 1                        & 1                        & 2.27                     & 2      & 2    & 3.70 & 4      & 4    & 2.36 & 2      & 3    \\
	                                 & van Gogh                & 1.51                     & 1                        & 1                        & 3.49                     & 4      & 4    & 2.86 & 3      & 3    & 2.14 & 2      & 2    \\
	\cmidrule{2-14}
	                                 & CATEGORY TOTAL          & 1.82                     & 1                        & 1                        & 2.52                     & 2      & 2    & 3.23 & 3      & 4    & 2.43 & 2      & 2    \\
	\midrule
	\multirow{4}{*}{\rotatebox{90}{Architecture}}
	                                 & Duomo-Taj Mahal         & 2.14                     & 2                        & 1                        & 2.20                     & 2      & 1    & 2.7  & 3      & 3    & 2.91 & 3      & 4    \\
	                                 & Eiffel-Sagrada Familia  & 1.95                     & 2                        & 1                        & 2.28                     & 2      & 3    & 3.42 & 4      & 4    & 2.35 & 2      & 2    \\
	                                 & Pisa-Big Ben            & 3.04                     & 3                        & 4                        & 2.13                     & 2      & 1    & 2.52 & 3      & 4    & 2.31 & 2      & 2    \\
	\cmidrule{2-14}
	                                 & CATEGORY TOTAL          & 2.38                     & 2                        & 1                        & 2.20                     & 2      & 1    & 2.90 & 3      & 4    & 2.52 & 2      & 2    \\
	\midrule
	\multicolumn{2}{c}{GLOBAL TOTAL} & 2.33                    & 2                        & 1                        & 2.32                     & 2                        & 2      & 2.75 & 3    & 4      & 2.60 & 3    & 2             \\
	\bottomrule
\end{tabular}

%% file: sections/conclusions.tex

This paper investigated how text-to-image diffusion models can blend two distinct concepts, introduced as separate prompts, into a single coherent image. We examined four zero-shot blending strategies, each leveraging a different aspect of the diffusion process, and showed that they can operate effectively without specialized training, fine-tuning, or complex prompt engineering. Although each strategy can produce high-quality hybrids, our results indicate that no single approach dominates in all scenarios; rather, blending outcomes depend heavily on the nature of the concept pair (e.g., animal–animal vs. architectural structures), the initial noise seed, and the user's subjective notion of a ``successful'' blend.
By systematically evaluating the proposed methods across five concept categories, we showed the zero-shot compositional capability of diffusion models. This capability underscores the models' ability to perform creative fusion of ideas, complementing broader AI research on compositional reasoning and generative creativity \cite{park2024rare}. Our user study with 100 participants provided clear empirical evidence of where each strategy excels, reinforcing that the iterative approaches (\SWI and \ALT) are often preferred but remain sensitive to user choices, such as prompt order and the point of switching, while \UNE and \PRO handle certain cases reliably but can falter when concepts are highly dissimilar or insufficiently anchored.
The findings presented here align with ongoing research in knowledge representation, interactive design, and computational creativity \cite{liu2022compositional}. This opens up creative applications, such as rapid design prototyping or cross-domain concept generation, that extend beyond conventional use cases like style transfer \cite{huang2023composer}.
We also discussed the limitations of our work. First, our study mostly focused on short, discrete prompts, thus leaving open questions about how these methods handle longer or more complex descriptions. Second, the inherent variability of diffusion outputs can lead to unpredictable or ``cartoonified'' images, especially when the model struggles with the newly introduced concept. Third, the user study sampled general participants; results might differ with domain experts or professional artists.

As future work, several concrete next steps arise from this study. We aim to test our four blending methods within more recent diffusion models, e.g., SDXL \cite{stable-diffusion-xl}, which could yield more refined or stable outputs. An interesting avenue is to incorporate symbolic constraints or partial blending into the pipeline, ensuring that only certain features of one concept transfer to the other. Multi-step blending (e.g., merging more than two concepts) may further stretch the boundaries of creative composition. Finally, large-scale user studies with participants from specialized domains, such as art, design, or psychology, could indicate additional nuances in aesthetic preferences and the cognitive interpretation of blended images.

Although we did not explicitly compare to fine-tuned or specialized style-transfer frameworks (e.g., DreamBooth-like methods \cite{ruiz2023dreambooth}), our focus on zero-shot methods fills a distinct niche: it shows how well diffusion models as is handle concept fusion. By showcasing practical, zero-shot blending techniques, we have highlighted diffusion models’ latent potential for creative synthesis without specialized training or prompt engineering. We encourage the community to build upon our publicly released codebase, prompts, and user-study protocols to replicate and extend our findings in more diverse application contexts.

%% file: sections/appendix.tex
\begin{figure}[t!]
	\centering
	\begin{overpic}
		[width=0.95\linewidth]{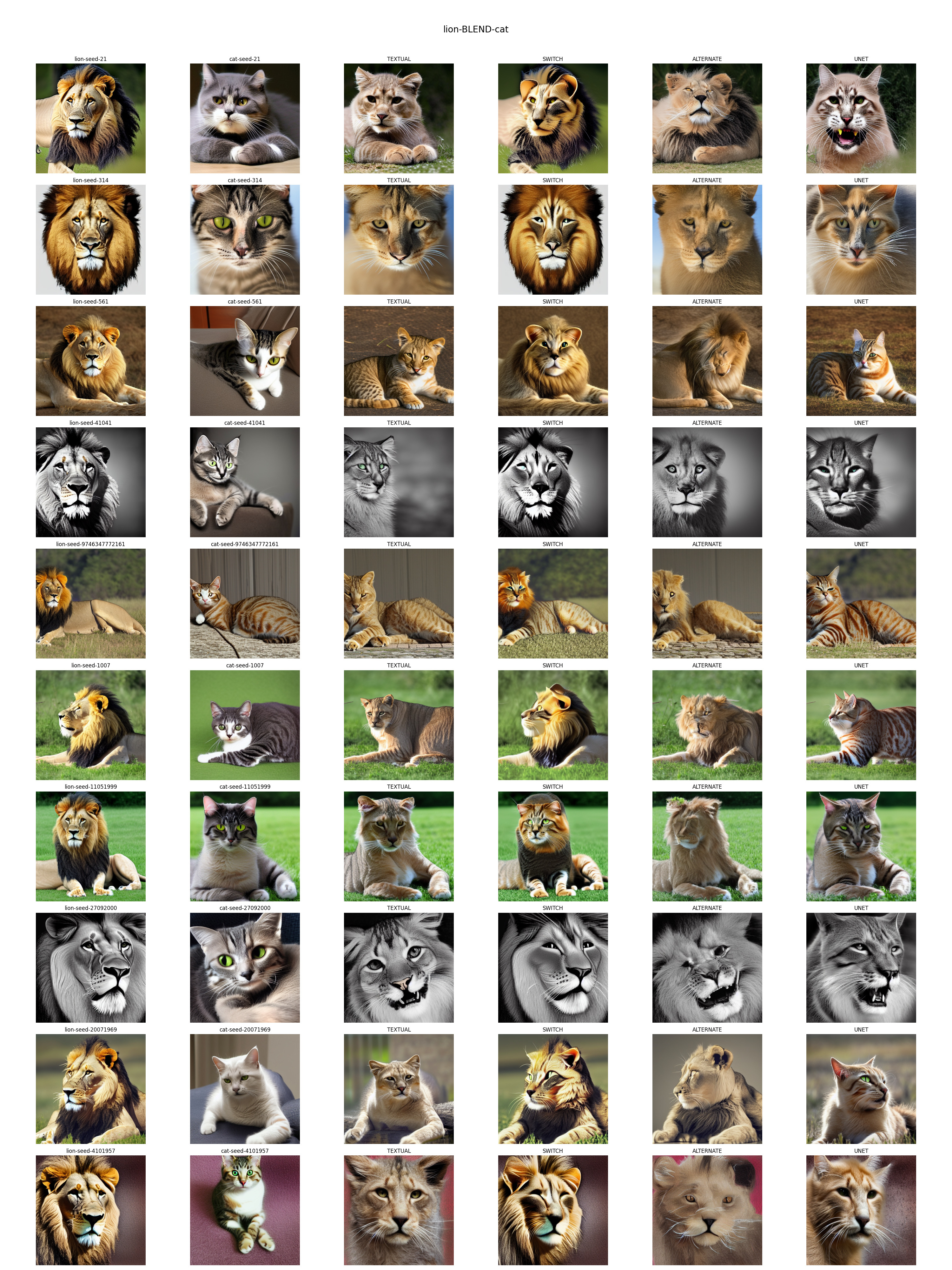}
	\end{overpic}
	\caption{
		Entire batch of $10$ images per blending methods generated by our
		implementation when blending the prompts ``\emph{lion}'' and
		``\emph{cat}''. The first two rows are the images generated respectively
		with the first and second prompt, the other are the blended images with
		each blending method.
	}
	\label{fig:full-lion-cat}
\end{figure}